\documentclass[conference, twocolumn]{IEEEtran}
\usepackage{listings}
\usepackage{graphicx}   
\usepackage{subcaption}
\usepackage[table,xcdraw]{xcolor}
\usepackage{amsmath}
\usepackage{pifont}
\usepackage{graphicx}
\usepackage{amssymb}
\definecolor{mydarkblue}{rgb}{0,0.08,0.65}
\usepackage[colorlinks=true,
    linkcolor=mydarkblue,
    citecolor=mydarkblue,
    filecolor=mydarkblue,
    urlcolor=mydarkblue]{hyperref} 
\usepackage{cleveref}
\usepackage{soul}
\usepackage[shortlabels]{enumitem}
\usepackage{url}
\usepackage{listings}
\usepackage{color}
\usepackage{tikz}
\usepackage{balance}
\usepackage{multirow}
\usepackage{comment}
\usepackage{amsmath}
\usepackage{graphicx}
\usepackage{wrapfig,lipsum,booktabs}
\usepackage{titlesec}
\usepackage[frozencache,cachedir=.]{minted}
\newcommand\numberthis{\addtocounter{equation}{1}\tag{\theequation}}
\usepackage[symbol]{footmisc}
\usepackage{tablefootnote}
\usepackage{tabularx}
\usepackage{placeins}

\usepackage{cuted}
\usepackage{float}
\usepackage{caption}

\usepackage{graphicx}
\usepackage{subcaption}
\usepackage{multicol}

\usepackage{dblfloatfix}
\usepackage{natbib}

\newcommand{\cmark}{\ding{52}}
\newcommand{\xmark}{\ding{53}}

\definecolor{codegreen}{rgb}{0,0.6,0}
\definecolor{codegray}{rgb}{0.5,0.5,0.5}
\definecolor{codepurple}{rgb}{0.58,0,0.82}
\definecolor{backcolour}{rgb}{0.95,0.95,0.92}

\makeatletter
\def\blfootnote{\xdef\@thefnmark{}\@footnotetext}
\makeatother

\lstdefinestyle{mystyle}{
  backgroundcolor=\color{backcolour},   commentstyle=\color{codegreen},
  keywordstyle=\color{magenta},
  numberstyle=\tiny\color{codegray},
  stringstyle=\color{codepurple},
  basicstyle=\ttfamily\footnotesize,
  breakatwhitespace=false,         
  breaklines=true,                 
  captionpos=b,                    
  keepspaces=true,                 
  numbers=left,                    
  numbersep=5pt,                  
  showspaces=false,                
  showstringspaces=false,
  showtabs=false,                  
  tabsize=2,
}

\AtBeginDocument{%
  \providecommand\BibTeX{{%
    \normalfont B\kern-0.5em{\scshape i\kern-0.25em b}\kern-0.8em\TeX}}}

\IEEEoverridecommandlockouts
\begin{document}

\title{Compressed Convolutional Attention: Efficient Attention in a Compressed Latent Space}

\author{Tomas Figliolia $\quad$ Nicholas Alonso  $\quad$ Rishi Iyer $\quad$ Quentin Anthony $\quad$ Beren Millidge \\
{
\small
\{tomas, nick, rishi, quentin, beren\}@zyphra.com
}\\
{}\\
{
 Zyphra
 \small
}\\
{
\small
 Palo Alto, CA
}
}

\maketitle

\setcounter{page}{1}

\begin{abstract}
Multi-headed Attention's (MHA) quadratic compute and linearly growing KV-cache make long-context transformers expensive to train and serve. Prior works such as Grouped Query Attention (GQA) and Multi-Latent Attention (MLA) shrink the cache, speeding decode, but leave compute, which determines prefill and training speed, largely unchanged. We introduce Compressed Convolutional Attention (CCA), a novel attention method which down-projects queries, keys, and values and performs the entire attention operation \emph{inside the shared latent space}. This simple design dramatically cuts parameters, KV-cache, and FLOPs all at once by the desired compression factor. 
Because CCA is orthogonal to head-sharing, we combine the two to form Compressed Convolutional Grouped Query Attention (CCGQA), which further tightens the compute–bandwidth Pareto frontier so that users can tune compression toward either FLOP or memory limits without sacrificing quality. 
Experiments show that CCGQA consistently outperforms both GQA and MLA at equal KV-cache compression on dense and MoE models. Additionally, we show that CCGQA outperforms all other attention methods on MoE models with half the KV-cache of GQA and MLA, achieving an 8x KV-cache compression with no drop in performance compared to standard MHA. CCA and CCGQA also dramatically reduce the FLOP cost of attention which leads to substantially faster training and prefill than existing methods.
On H100 GPUs, our fused CCA/CCGQA kernel reduces prefill latency by about $1.7\times$ at a sequence length of 16k relative to MHA, and accelerates backward by about $1.3\times$.
\end{abstract}



\section{Introduction}

Self-attention is the core sequence-mixing component of the ubiquitous transformer architecture \citep{vaswani2017attention,gpt3,kaplan2020scaling}. While self-attention is extremely expressive, enabling every token to attend to every other token, this expressivity comes at a high computational cost. Attention has quadratic complexity in both the model hidden dimension and sequence dimension. Moreover, during autoregressive generation the KV-cache grows linearly with sequence length and hidden size, incurring huge memory-bandwidth costs that cannot be amortized across batches. These issues become especially problematic with long contexts in "reasoning models" \citep{guo2025deepseek,lambert2024t}, which produce extended chains of thought before responding. Serving models with KV-caches larger than a single GPU requires expensive context-parallelism approaches like Ring or Tree Attention \citep{liu2023ring,shyam2024tree} and makes autoregressive decoding memory-bandwidth bound, limiting hardware utilization.

A number of approaches in the literature have attempted to address this fundamental bottleneck. One popular approach is to create novel architectures such as state-space-models that eliminate the quadratic nature of self-attention entirely and instead utilize a constant size state rather than a linearly growing KV-cache \citep{gu2021efficiently,gu2023mamba,katharopoulos2020transformers,yang2024gla,sun2023retentive,peng2023rwkv}. However, these architectures tend to be less expressive than attention and often underperform on more complex tasks requiring sustained reasoning or in-context learning \cite{jelassi2024repeat, park2024mamba,grazzi2024mamba}. This led to the proposal of hybrid architectures \citep{glorioso2024zamba,glorioso2024zamba2,lieber2024jamba,waleffe2024empirical,Bamba}, which combine both SSMs and attention, and can achieve the best of both worlds. However, since hybrids still contain some attention, they do not ultimately eliminate the quadratic bottleneck either. Other approaches maintain the quadratic nature of self-attention but try to compress the KV-cache directly, usually in an offline fashion \citep{ge2023context,yang2024lossless,kim2024lexico,liu2024clusterkv}. Such methods can achieve significant compression rates, but often come at a large cost to generation quality. 

While these methods are more speculative and have not been ubiquitously adopted, fundamental improvements have also been made in the core self-attention operation. Multi-query attention (MQA) \citep{shazeer2019fast}, and grouped query attention (GQA) \citep{ainslie2023gqa} keep the core self-attention primitive and reduce the KV-cache requirements by parameter-sharing across K and V heads. GQA has seen significant recent adoption for language model pretraining being used in \cite{jiang2023mistral,touvron2023llama,abdin2024phi} and results in significant improvement in inference speed. Recently, another attention variant named Multi-Latent Attention (MLA) \citep{deepseekv3} took a different approach by learning to directly compress the K and V projections needed for the KV-cache. 

We consider GQA and MLA to each be instances of separate strategies for reducing the KV-cache. GQA uses parameter-sharing of the K and V heads. This reduces memory requirements, since the KV-cache effectively consists of many copies of the same K and V heads, which can be merged. GQA does not reduce the FLOPs needed for training or prefill compared to MHA. However, for autoregressive generation, which is typically memory-bandwidth-bound, the savings in parameters that need to be loaded per token leads to a substantial increase in throughput.

MLA, by contrast, compresses the keys and values into
a smaller, learnable, and shared subspace, which can then
be used for generation. It uses this learnt subspace for storage of the KV-cache and up-projects back to the full dimension for the actual attention operation. This means that MLA does not offer compute savings in training or prefill over MHA or GQA and is in fact slightly more expensive in compute and parameters due to the up- projections. MLA also has additional complications due to
RoPE \citep{su2023rotary}, which cannot operate directly on the
compressed cache, so MLA must keep a separate key RoPE
cache shared across heads. During decoding, however, MLA
possesses an ‘MQA-mode’ which merges the up-projections for the shared-KV cache into the query up-projection and output projection.
This significantly reduces the memory bandwidth-required during decoding.

Both GQA and MLA focus primarily on reducing the KV-cache, which is important for decoding speed, but
do not meaningfully reduce the fundamental compute cost of
attention which is the performance bottleneck in both training
and inference prefill. Prefill performance is especially impor-
tant during long-context workloads where the vast majority of
tokens are inputs to the model rather than generated.

In this paper, we present Compressed Convolutional Attention (CCA), an elegant parameter- and compute- compression method, which removes the drawbacks of MLA and outperforms it in practice. Specifically, CCA performs the attention operation \emph{entirely in the compressed latent space}, meaning that it can reduce the compute required for attention by a factor of the query compression rate, but also that it can seamlessly integrate RoPE without requiring separate heads and cache. By performing the full attention in the compressed space, CCA has substantial parameter savings at a fixed compression rate compared to MLA since it eschews the needs for QKV up-projection matrices. 

While we found that naively performing attention on the compressed QKV latents incurs a significant performance penalty, we discovered that by performing additional convolutional sequence and channel mixing on the compressed Q and K latents, we can exceed the performance of MLA and even MHA. CCA outperforms all other methods, including both GQA and MLA at equal KV-cache compression rates in MoE settings with 4x less FLOPs.

Moreover, since parameter-compression and parameter-sharing approaches can be jointly utilized, we find it is productive to combine them in a method we call Compressed Convolutional Grouped Query Attention (CCGQA) which applies a GQA-style K and V head sharing within the already compressed latent space. This enables us to achieve an additional 2x KV cache reduction without performance penalty. (CCGQA also allows us to decouple the compression rates of queries and keys, since we can replicate compressed keys to match less compressed queries). Additionally, we show that in a parameter-matched setting for MoEs, CCA can outperform MLA along both compute and KV-cache dimensions, thus creating a smooth Pareto frontier for CCA and CCGQA in which a trade-off can be made between compute-bound and memory-bandwidth-bound scenarios without sacrificing performance. Thus, CCA is a highly versatile method capable of improving performance in both single- or large-batch inference settings and can be flexibly adjusted to a wide variety of possible parallelism settings during pretraining.

In addition to the benefits in model quality at a fixed parameter budget and memory overhead, executing attention entirely in the compressed latent makes CCA—and its grouped variant CCGQA—substantially faster in practice. Because the \(S^2\) terms in \(QK^\top\) and \(\mathrm{Attn}\!\cdot\!V\) shrink by \(1/C\), the speedup compared to MHA grows with sequence length. We implemented a fused kernel for CCGQA and found that on an H100 in BF16 with \(E{=}2048\), CCA-4\(\times\) yields \(\approx 1.6\!-\!1.7\times\) lower prefill latency than MHA at \(S{=}16\text{k}\) across head dimensions \(d_h\!\in\!\{64,128,256\}\) (\(\sim1.3\!-\!1.4\times\) vs.\ GQA-8 and \(\sim1.3\!-\!1.5\times\) vs.\ MLA), while forward-causal shows \(\sim1.6\!-\!1.9\times\) gains. Training backward is also faster by \(\approx 1.2\!-\!1.3\times\) at the same length. A decoupled CCA configuration with \(C_1{=}2\) for queries and \(C_2{=}8\) for keys/values retains strong efficiency, delivering \(\sim1.3\!-\!1.4\times\) prefill speedup over MHA and \(\sim1.1\!-\!1.3\times\) over GQA-8 at \(S{=}16\text{k}\), while preserving the KV-cache savings and the theoretical \(1/C\) scaling.

\begin{table}[htbp]
\centering
\caption{Notation used throughout the paper.}
\begin{tabularx}{\linewidth}{@{}cX@{}}
\toprule
\textbf{Symbol} & \textbf{Definition / role} \\
\midrule
$B$               & Batch size (number of sequences processed in parallel) \\[2pt]
$S$               & Sequence length (tokens per sequence) \\[2pt]
$E$               & Model embedding / residual dimension \\[2pt]
$n_h$             & Number of attention heads \\[2pt]
$d$               & Per-head dimension ($d = E / n_h$) \\[2pt]
$G$               & Number of groups in GQA \\[2pt]
$C$               & Compression factor in CCA ($C = E / \tilde e$) \\[2pt]
$C_{1},\,C_{2}$   & Query-compression and KV-compression factors in CCGQA \\[2pt]
$c_q$             & Query compression factor in MLA \\[2pt]
$c_{kv}$          & Key/Value compression factor in MLA \\[2pt]
$\tilde e$        & Latent dimension after compression ($\tilde e = E / C$) \\[2pt]
$d_h$             & Latent per-head size ($d_h = \tilde e / n_h$) \\[2pt]
$x\!\in\!\mathbb{R}^{S\times E}$ & Input/residual token embeddings \\[2pt]
$W_Q,W_K,W_V,W_O$ & Standard projection matrices ($E \!\times\! E$) \\[2pt]
$\tilde W_Q,\tilde W_K,\tilde W_V$ & Down-projections to latent space ($E \!\times\! \tilde e$) \\[2pt]
$C_{KV}$          & Shared compressed key-value cache \\[2pt]
$C_Q$             & Compressed query latent \\[2pt]
$\tilde q,\tilde k,\tilde v$ & Compressed query, key, and value tensors \\[2pt]
$o_h$             & Output of attention head $h$ \\[2pt]
$\beta$           & Learnable temperature scaling for keys \\[2pt]
\bottomrule
\end{tabularx}
\label{tab:notation}
\end{table}

\section{Method}

\begin{figure*}
    \centering
    \includegraphics[width=0.7\linewidth]{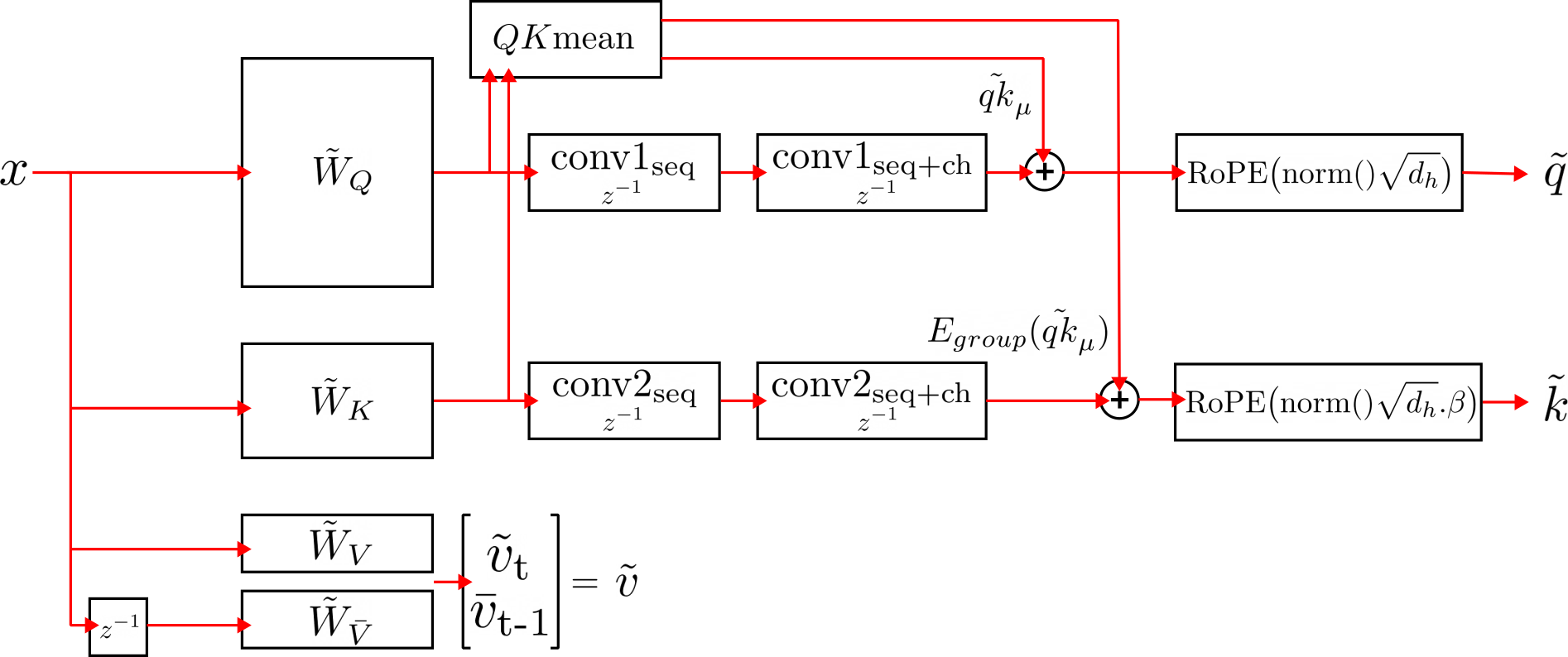}
    \caption{Diagram of the operations involved in the CCA block. This diagram describes the computation of the compressed latent query, key, and value vectors prior to performing standard Flash Attention on the compressed latents. The input $x$ is first down-projected using the $\tilde{W}_Q, \tilde{W}_K, \tilde{W}_V$ matrices, then the two convolution operations are performed followed by the QK-mean operation, then normalization. For the V matrix, we do not apply any convolutions, but instead apply the v-shift operation.}
    \label{fig:placeholder}
\end{figure*}

\subsection{Preliminaries}

\subsubsection{Multihead Attention (MHA)}

Classical multi-head attention (MHA)  \citep{vaswani2017attention} operates on the input embedding $x \in \mathcal{R}^{S \times E}$ coming from the residual stream in a transformer, where $S$ is the sequence length and $E$ is the residual stream dimension (we omit the batch dimension to keep notation clearer), and produces the query $q$, key $k$, and value $v$ matrices through separate linear projections of the embeddings:
\begin{align*}
    q &= x W_Q  \\
    k &= x W_K  \\
    v &= x W_V  \numberthis
\end{align*}
The $q ,k, v$ projections are split into separate independent `heads', such that $q = [q_1, q_2 \dots q_{n_h}]$ where $n_h$ is the number of heads. Each head has dimension $d =E/n_h$. The intuition behind this is that each head can be operated on independently and in parallel by attention, allowing them to specialize in attending to different aspects of the input.

Given the query, key, and value heads, we perform the softmax attention operation for each head independently then recombine the heads with the output projection $W_O$,
\begin{align*}
    o_h &= \text{softmax}\left(\frac{q_h k_h^T}{\sqrt{d}}\right) v_h \\
    \text{out} &= W_o [o_1, o_2 \dots o_{n_h}] \numberthis
\end{align*}
where $h$ is the head index. Due to the $q_h k_h^T$ term, attention computation scales quadratically in the sequence dimension. Since the $W_q, W_k, W_v, W_o$ matrices are of dimension $E \times E$, it also scales quadratically in the channel dimension. For autoregressive generation, we typically store each $k_h$ and $v_h$ for each sequence element $t$ in the KV-cache, which is thus of size $2 \times S \times E$.

\vspace{3pt}

\subsubsection{Grouped Query Attention (GQA)}

\begin{table*}[htbp]
\centering
\resizebox{\textwidth}{!}{
\setlength{\tabcolsep}{4pt}
\begin{tabular}{@{}lllll@{}}
\toprule
\textbf{Attention} & \textbf{Parameters} & \textbf{KV-cache} & \textbf{Forward FLOPs} & \textbf{Decode FLOPs} \\
\midrule
MHA & \(4E^{2}\) &
\(2BSE\) &
\(8BSE^{2} + 4BES^{2}\)  &
\(8BE^{2} + 4BES\) \\[2pt]
GQA &
\(2E^{2} + \!2\frac{E^2}{G}\) &
\(2\frac{BSE}{G}\) &
\((1+\frac{1}{G})(4BSE^{2}) + 4BES^{2}\)  &
\((1+\frac{1}{G})(4BE^{2}) + 4BES\) \\[2pt]
MLA &
\(E^{2} + 3\frac{E^{2}}{c_{kv}} +2\frac{E^{2}}{c_q} \) &
\(\frac{BSE}{c_{kv}} + BSE_{r}\) &
\(2BSE^{2}+4BS\frac{E^2}{c_q}+6BS\frac{E^2}{c_{kv}}  + 4BES^{2}\) & 
\((2BE^2+2BN_h{E^2})(\frac{1}{c_{kv}})+\frac{2BN_h{E^2}}{c_qc_{kv}}+\frac{2BE^2}{c_q}+ \frac{4BESN_h}{c_{kv}}\) \\[2pt]
CCA &
\(4\frac{E^{2}}{C} + \text{Conv}\) &
\(2\frac{BSE}{C}\) &
\((\frac{2}{C})(4BSE^{2}) + \frac{4BES^{2}}{C}+ \text{Conv}\) &
\((\frac{2}{C})(4BE^{2}) + \frac{4BES}{C}+ \text{Conv}\) \\[2pt]

CCGQA &
\(2\frac{E^{2}}{C_1} +2\frac{E^{2}}{C_2} + \text{Conv}\) &
\(2\frac{BSE}{C_2}\) &
\((\frac{1}{C_1}+\frac{1}{C_2})(4BSE^{2}) + \frac{4BES^{2}}{C_1}+ \text{Conv}\) &
\((\frac{1}{C_1}+\frac{1}{C_2})(4BE^{2}) + \frac{4BES}{C_1}+ \text{Conv}\)\\[2pt]
\bottomrule
\multicolumn{5}{p{\linewidth}}{\scriptsize%
$\dagger$\,\textbf{Conv term.}  For CCA let $\tilde e = E/C$, $h$ be the number of heads.
Conv‐layer parameters $= 2\tilde e\,k_{\mathrm{seq}}
           + \dfrac{\tilde e^{2}}{h}\,k_{\mathrm{ch}}$. 
Training FLOPs $= 2BS\!\left(2\tilde e\,k_{\mathrm{seq}}
           + \dfrac{\tilde e^{2}}{h}\,k_{\mathrm{ch}}\right)$;
decode FLOPs  $= 2B\!\left(2\tilde e\,k_{\mathrm{seq}}
           + \dfrac{\tilde e^{2}}{h}\,k_{\mathrm{ch}}\right)$.
For CCGQA replace $\tilde e$ by 
$\bar e=\dfrac{E}{C_1}+\dfrac{E}{C_2}$ and
$h$ by $h_q+h_k$.  Kernel sizes are
$k_{\mathrm{seq}}$ (depth-wise causal) and $k_{\mathrm{ch}}$ (grouped).}
\end{tabular}
}
\caption{Comparison of parameter count, KV-cache size, and computational complexity (FLOPs) for different attention mechanisms. $E$ denotes the residual embedding dimension, $B$ is the batch size, $S$ is the sequence length, $G$ is the group size in GQA and $C$ denotes the compression factor in CCA. For MLA we use separate compression factors $c_q$, $c_{kv}$ to denote the low rank dimensions for query (with RoPE) and the shared KV-cache (we omit the shared key-RoPE and query-RoPE inference projection for the FLOPs calculation. This means that MLA's FLOPs are actually slightly higher than shown). For CCGQA we introduce two separate compression factors, one for queries and one for keys and values, denoted by $C_1$, $C_2$. $E_r$ denotes the RoPE dimension in MLA. We observe that CCA has both significantly fewer parameters and smaller KV-cache requirements than alternative methods and also, unlike all other methods, reduces the ultimate attention FLOPs required. A full table of notation can be found in Table \ref{tab:notation}}
\label{tab:attention_comparison}
\end{table*}

\vspace{3pt}

GQA \citep{ainslie2023gqa} is a method that aims to reduce the KV-cache size by sharing the parameters across K and V heads. Specifically, GQA splits the heads up into groups and sets the parameters of all the heads in a group to be equal. For example, for two groups of four heads, GQA asserts the following equality: $k_{g1} = k_1 = k_2 = k_3 = k_4; k_{g2} = k_5 = k_6 = k_7 = k_8$. Since all the heads within a group share the same parameters, they do not need to be stored separately within the KV-cache, allowing us to achieve $n_h/G$ memory savings, where $G$ is the number of groups. MQA \citep{shazeer2019fast} is the extremal version of GQA with $G = 1$, such that all pairs of key and value heads share the same parameters which reduces the KV-cache memory by a factor of $n_h$. However, this typically comes at a steep performance cost. GQA lets us interpolate between MHA and MQA to smoothly trade-off KV-cache size and model quality. 

\begin{figure*}[!htbp]
  \centering
  \begin{subfigure}[b]{0.48\textwidth}
    \centering
    \includegraphics[width=\textwidth]{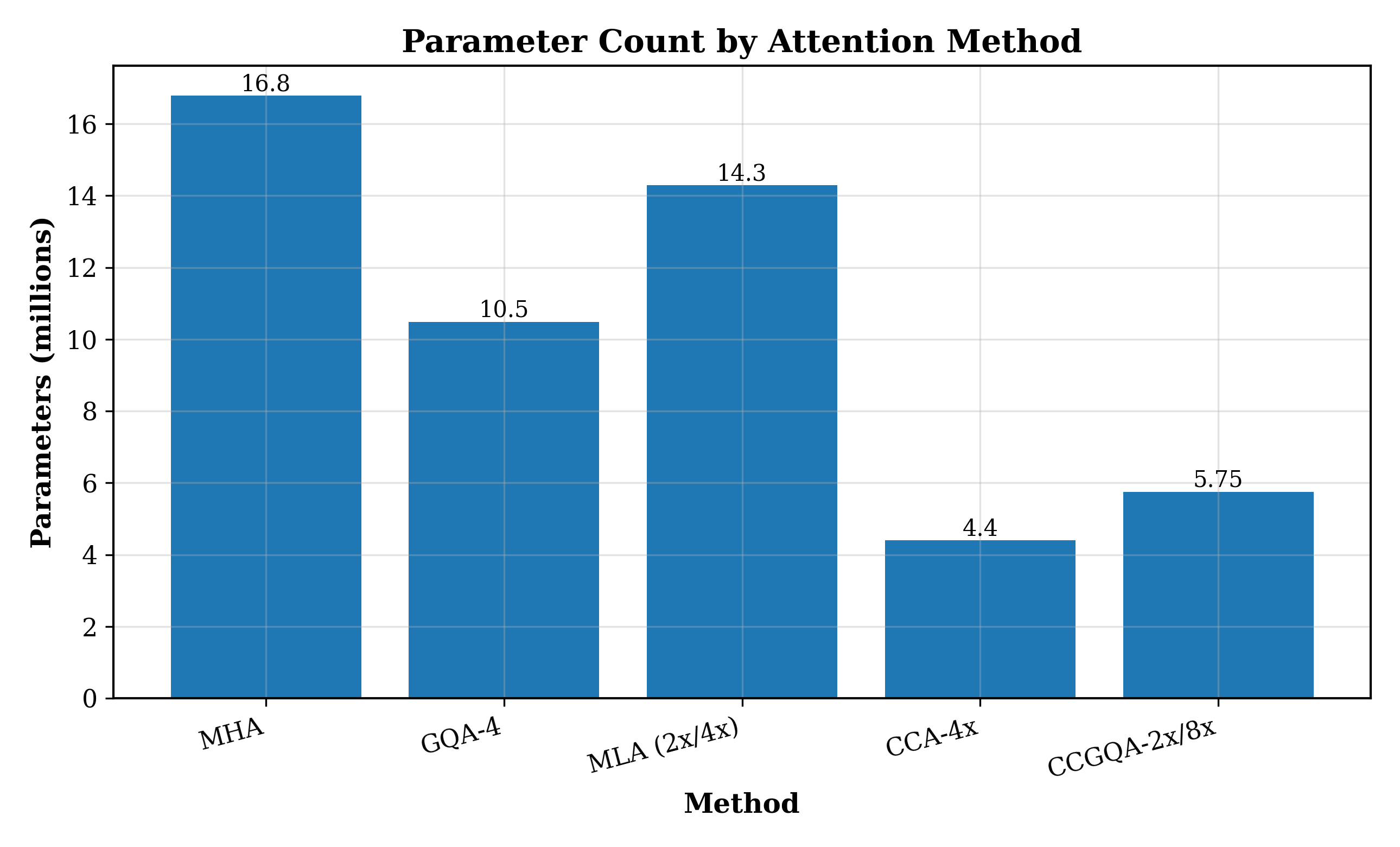}
    \caption{Parameter count comparison}
    \label{fig:params-bar}
  \end{subfigure}
  \hfill
  \begin{subfigure}[b]{0.48\textwidth}
    \centering
    \includegraphics[width=\textwidth]{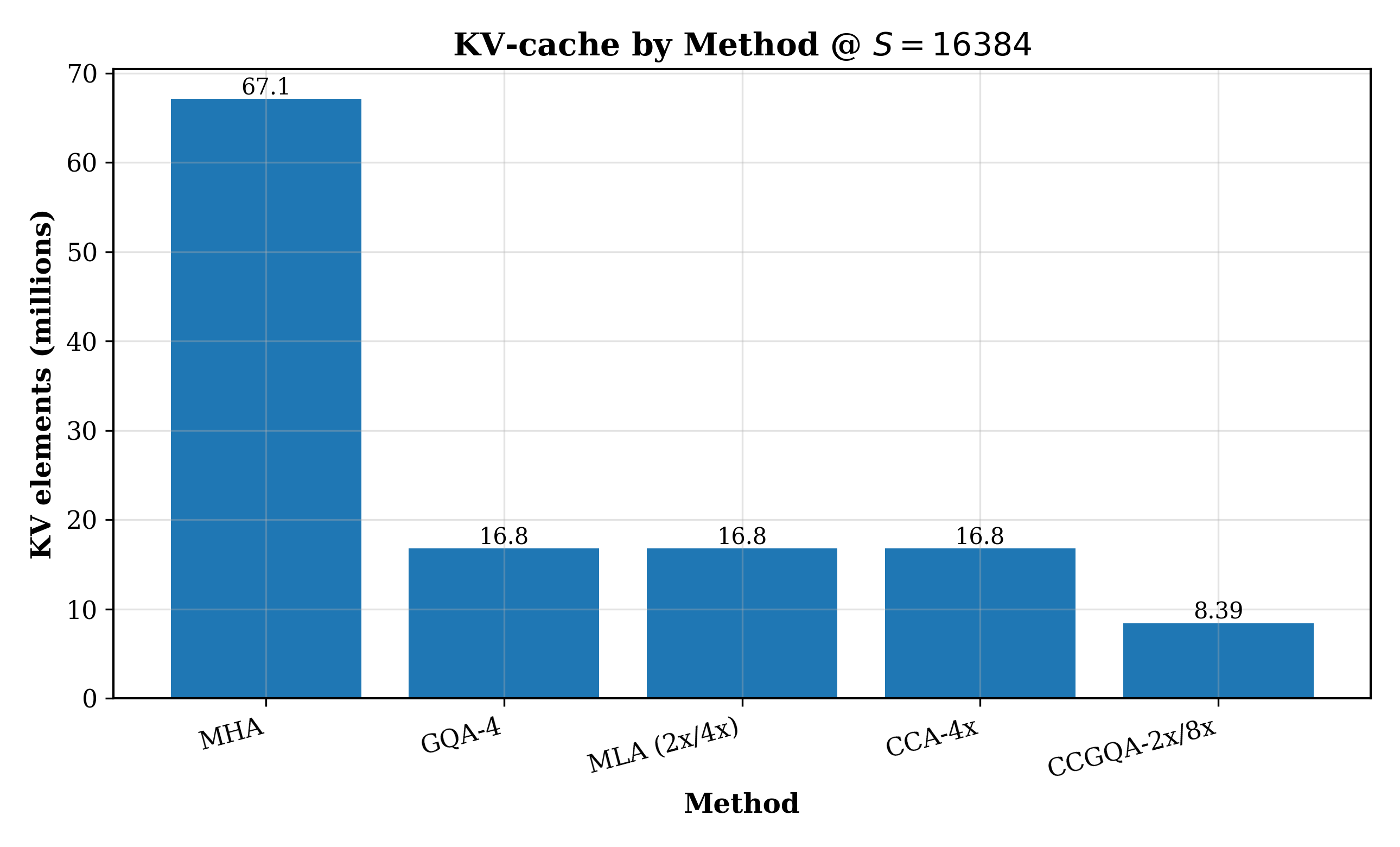}
    \caption{KV-cache at $S=16384$}
    \label{fig:kv-bar-slong}
  \end{subfigure}
  
  \vspace{0.5em}
  
  \begin{subfigure}[b]{0.48\textwidth}
    \centering
    \includegraphics[width=\textwidth]{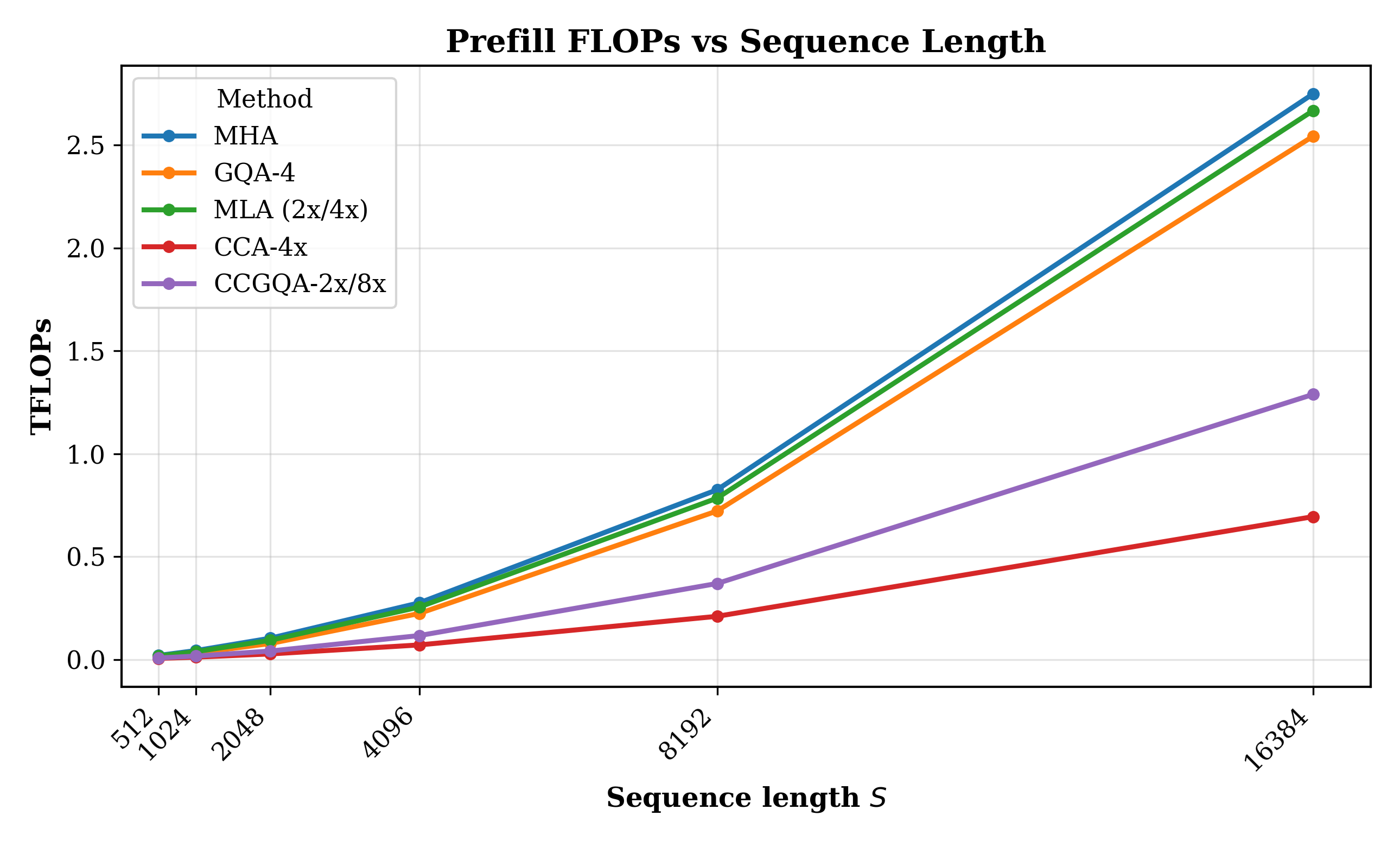}
    \caption{Prefill FLOPs vs sequence length}
    \label{fig:fwd-line}
  \end{subfigure}
  \hfill
  \begin{subfigure}[b]{0.48\textwidth}
    \centering
    \includegraphics[width=\textwidth]{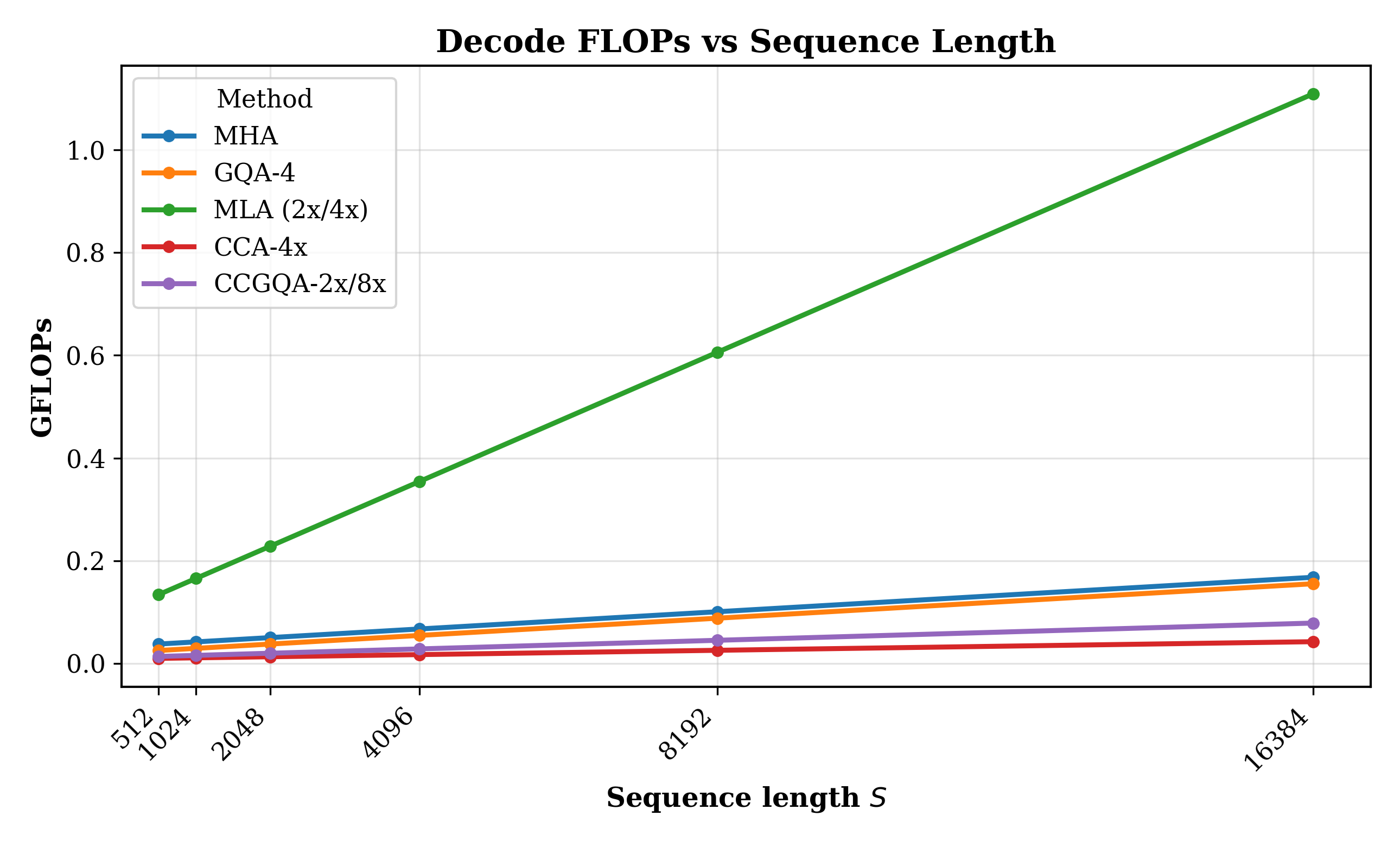}
    \caption{Decode FLOPs vs sequence length}
    \label{fig:dec-line}
  \end{subfigure}
  
  \caption{Theoretical computational and memory complexity analysis across attention mechanisms with $E=2048$. (a) Parameter counts show compression methods reduce model size. (b) KV-cache memory at long context demonstrates substantial savings. (c) Prefill and (d) decode FLOPs exhibit quadratic and linear scaling with sequence length respectively. Note that these are theoretical FLOP counts. See empirical latency measurements in Figures~\ref{fig:fwd-nocausal-64}-\ref{fig:bwd-256}. Our kernel implementation will be further improved through better operator fusion and memory access patterns. See Appendix \ref{sec:appendix-mla-inference} for more details on MLA inference considerations.}
  \label{fig:complexity-analysis}
\end{figure*}

\vspace{3pt}

\subsubsection{Multi-Latent Attention (MLA)}

\vspace{3pt}

MLA takes a different approach to reducing KV-cache size. Instead of sharing parameters, MLA projects the K and V parameters into a smaller shared latent space $C_{KV}\in \mathcal{R}^{S \times \tilde{e}}$ from which the full K and V can then be (lossily) reconstructed. Here, $\tilde{e}$ is the dimension of the compressed space. The compression is done with linear down-projections, such as $\tilde{W}_{DKV} \in \mathcal{R}^{E \times \tilde{e}}$,
\begin{align*}
    C_{KV} &= x \tilde{W}_{DKV}  \\
    C_Q &= x \tilde{W}_{DQ}  \numberthis
\end{align*}
In the KV-cache, we then only need to store the compressed representation $C_{KV}$ which is $\tilde{e}/E$ times smaller than the uncompressed MHA KV-cache. In MLA we typically compress queries by a smaller factor than the keys and values. Given the compressed cache, the K and V heads are generated using separate up-projection matrices of shape $\tilde{e} \times E$,
\begin{align*}
    [q_1,q_2 \dots q_{n_h}] &= \tilde{W}_{UQ} C_{Q} \\
    [k_1,k_2 \dots k_{n_h}] &= \tilde{W}_{UK} C_{KV} \\
    [v_1, v_2 \dots v_{n_h}] &= \tilde{W}_{UV} C_{KV} \numberthis
\end{align*}

In the NoPE \citep{wang2024lengthgeneralizationcausaltransformers} setting, MLA is advantageous as one can amortize all up-projections at test time and transform into MQA using the following relation:

\begin{align*}
    \mathrm{softmax}(\underbrace{C_Q^\top {W}_{UQ}}_{q \in \mathbb{R}^{1 \times d_h}} \underbrace{{W}_{UK}^\top C_{kv}}_{k^\top \in \mathbb{R}^{d_h \times N}}) \underbrace{C_{kv}^\top {W}_{UV}}_{v \in \mathbb{R}^{N \times d_h}} W^o = \\\mathrm{softmax}(\underbrace{C_Q^\top {W}_{UQ} {W}_{UK}^\top}_{q' \in \mathbb{R}^{n_h \times d}} \underbrace{C_{kv}}_{k'^\top \in \mathbb{R}^{d \times N}}) \underbrace{C_{kv}^\top}_{v' \in \mathbb{R}^{N \times d}} {W}_{UV} W^o \numberthis
\end{align*}
The case of RoPE, however, is more complicated. We cannot apply the RoPE rotation matrix directly to the $C_{KV}$ cache if we want to merge up- and down-projections perfectly at inference time. In MLA, the key RoPE head is shared across all heads. To handle this, we define special RoPE heads and RoPE cache,
\begin{align*}
    q^r &= \text{RoPE}(W_{UQR} C_Q) \\
    k^r &= \text{RoPE}(W_{KR} x) \\
    q &= [q_1, \dots q_{n_h}, q^r_1 \dots q^r_{n_{h}}] \\ 
    k &= [k_1, \dots k_{n_h}, k^r \dots k^r_{n_{h}}]  \numberthis
\end{align*}

With the concatenated RoPE and non-RoPE query and key heads, attention is then performed as in MHA. By compressing K and V into the shared latent $C_{KV}$, MLA can achieve significant reductions in KV-cache memory and attention parameters vs MHA. However, the up-projections add significant parameter overhead, and the shared key RoPE decreases expressivity of the positionally encoded information. In addition, since the q, k, v vectors are up-projected prior to attention, MLA uses approximately the same number of FLOPs in attention as MHA during training. During inference, MLA is advantageous due to the conversion of MHA to MQA via the amortized up-projections when memory-bandwidth bound (see Appendix B for a detailed discussion). While being parameter efficient, MQA is not entirely advantageous in multi-device inference in tensor-parallel settings, as each device would receive a copy of the shared KV latent before performing attention, which wastes potential expressivity.

\subsection{Compressed Convolutional Attention (CCA)}

Our proposed CCA method follows MLA and GQA in compressing the KV-cache, but performs attention purely in the compressed latent space. This enables RoPE to be seamlessly integrated without the need of special RoPE heads and projections, reduces the training parameter cost by more than $2\times$ compared to MLA since we no longer require up-projection matrices, and reduces attention FLOPs by the compression factor $E/\tilde{e}$, which can be substantial even if it does not directly address the quadratic compute bottleneck. For instance, CCA with a $16\times$ compression enables $\sqrt{16}=4\times$ longer sequences to be processed for the same FLOP budget. Like MLA, CCA performs a linear down-projection of the queries, keys, and values into a compressed latent space. However, to maintain and enhance performance, CCA then performs three key innovations: convolutional mixing across both sequence and channel dimensions, q-k-mean, and value-shift prior to performing standard attention on the adjusted q, k, v. 

The first step of CCA is projecting q and k into the compressed latent space with matrices $\tilde{W}_Q, \tilde{W}_K \in \mathcal{R}^{E \times \tilde{e}}$:
\begin{align*}
    \tilde{q} &= [\tilde{q_1}, \tilde{q}_2, \dots \tilde{q}_{n_h}] = x\tilde{W}_Q \\
    \tilde{k} &= [\tilde{k_1}, \tilde{k}_2, \dots \tilde{k}_{n_h}] = x\tilde{W}_K \numberthis
\end{align*}
Unlike MLA, since we are performing the full attention in the compressed latent space, we compress the queries by the same factor as keys and values. In the CCGQA case, where we are repeating key and value heads, we can compress queries less, up to a multiple of the number of groups. Concretely, CCGQA uses separate projection matrices $\tilde{W}_Q \in \mathcal{R}^{E \times E/C_1}$ and $\tilde{W}_K \in \mathcal{R}^{E \times E/C_2}$ with $C_2 \geq C_1$, so the key projection produces a smaller latent than the query projection.

We find that performing convolutional mixing across both the sequence and channel dimensions (within a head) for q and k can substantially improve the resulting performance of CCA. Our intuition is that these convolutions give additional expressivity to the transforms learned in the latent space, and that this additional smoothing allows better information transfer and preservation through attention, similar to how the causal convolution prior to the SSM in Mamba \citep{gu2023mamba} improves sequence mixing performance. In a similar vein, recently-proposed ``canon layers'' are convolutional mixing layers applied across MLP and attention layers \citep{Zhu2025canon}. Unlike canon layers, we only apply convolutions to q and k operations within the attention blocks, and use a sequence of two convolution layers instead of one. We find that mixing both across channels within a head (ch) and across the sequence (seq) is helpful.
\begin{align*}
    \tilde{q} &= \text{conv2}_{\text{seq+ch}}\big(\text{conv1}_{\text{seq}}(\tilde{q})\big) \\
    \tilde{k} &= \text{conv2}_{\text{seq+ch}}\big(\text{conv1}_{\text{seq}}(\tilde{k})\big) \numberthis
\end{align*}

\begin{table*}[htbp]
\centering
\begin{tabular}{|c|c|c c c c c c|}
\hline
\multicolumn{8}{|c|}{} \\
\hline
\textbf{Model} & Loss  & HellaSwag & ARC Easy & ARC HARD & Piqa & Winogrande & Avg \\
\hline
MHA & 2.297 & 58.9 & 63.4 & 37.2 & 74.5 & 56.8 & 58.2 \\
MLA & 2.321 & 57.8 & 63.3 & 35.4 & 74.6 & 57.6 & 58.2\\
GQA & 2.297 & 58.6 & 62.3 & 34.7 & 73.5 & 56.9 & 57.2\\
CCA & 2.307 & 57.4 & 62.4 & 34.7 & 74.0 & 56.7 & 57.0\\ 
CCGQA & 2.286 & 59.6 & 62.6 & 36.0 & 75.0 & 59.7 & 58.6 \\
\hline
\end{tabular}
\caption{Table of loss and evaluation scores for CCA, MHA, and related attention methods for a dense 1B parameter model trained for 300B tokens on the Zyda2 dataset. }
\end{table*}

Next, we perform the q-k-mean operation, which adds the mean of the values of q and k pre and post convolution to the convolved values. Intuitively, this helps share information between q and k, and also allows the model to interpolate the strength of the convolution by providing a skip connection. Geometrically, this increases the sparsity of the attention diagonal when combined with QK-norm. We compute this as follows:
\begin{align*}
    \tilde{qk}_\mu &= \frac{1}{2}(\tilde{q}_{\text{pre}} + B_{\text{group}}(\tilde{k}_{\text{pre}})) \\
    \tilde{q} &= \tilde{q} + \tilde{qk}_\mu \\
    \tilde{k} &= \tilde{k} + E_{\text{group}}(\tilde{qk}_\mu) \numberthis
\end{align*}
Where $\tilde{q}_{\text{pre}}$ and $\tilde{k}_{\text{pre}}$ are the q and k latent priors to the convolution. Operation $B_{\text{group}}(\cdot)$ is performing broadcast among the heads for the queries that have the same key assigned. Operation $E_{\text{group}}(\cdot)$ is just performing the mean across the heads that are combined into a single one for the keys. Both $B_{\text{group}}$ and $E_{\text{group}}$ are used in our GQA version of CCA, in which keys and values are shared in a group.

Finally, for the value projection, we introduce an operation referred to as \textit{value-shift}. Here, each attention head receives two distinct types of value vectors: one derived from the current input embedding, and another generated from the previous embedding in the sequence. These two value types are produced through independent transformations, each with its own set of parameters. Each one of these will generate the values needed for half of the heads.
\begin{align*}
    \tilde{v}_\text{t} &= \tilde{W}_V x_{t} \\
    \bar{v}_\text{t-1} &= \tilde{W}_{\bar{V}} x_{t-1} \\
    \tilde{v} &= [\tilde{v}_{\text{t}}, \bar{v}_{\text{t-1}}] \numberthis
\end{align*}
Given the compressed latent vectors $\tilde{q}, \tilde{k}, \tilde{v}$, we then perform Q, K L2 normalization and scale by the square-root of the head dimension, multiply the key by a learnable temperature parameter $\beta$ and apply RoPE before performing standard attention,
\begin{align*}
    \tilde{q} &= \text{RoPE}\big(\text{norm}(\tilde{q})\sqrt{d_h}\big) \\
    \tilde{k} &= \text{RoPE}\big(\text{norm}(\tilde{k})\sqrt{d_h} \cdot \beta \big) \\
    \tilde{o}_h &= \text{softmax}(\frac{1}{\sqrt{d}} \tilde{q}_h \tilde{k}_h^T) \tilde{v}_h \\
    \text{out} &= \tilde{W}_O [\tilde{o}_1, \tilde{o}_2 \dots \tilde{o}_{n_h}] \numberthis
\end{align*}
Where $\tilde{W}_O \in \mathcal{R}^{\tilde{e} \times E}$ performs the up-projection back to the dimension of the residual stream. While CCA may look somewhat complicated, the additional operations of the convolutions, q-k-mean-adjustment and value-shift require very few additional parameters and FLOPs while giving substantial improvements in perplexity and stability, which make performing the attention in the fully compressed latent space viable. 

CCA can also be combined with a GQA-style parameter sharing approach. Specifically, we can apply GQA directly to the compressed heads in the latent space. We apply GQA directly to the compressed key and value heads of CCA e.g. we share $\tilde{k}_{g1} = \tilde{k}_1 = \tilde{k}_2 = \tilde{k}_3 = \tilde{k}_4$ in the case where the group size is 4. We call this method Compressed Convolutional Group Query Attention (CCGQA). We find that this approach can again obtain the best-of-both-worlds of parameter sharing and compression, enabling us to achieve higher compression ratios for the same performance. Example pytorch code implementing CCA and CCGQA is provided in Appendix A.

\subsection{Methodology}
For our experiments, we do not do FLOPs/byte matched ablations. We instead opt for parameter-matched (total and active parameters) and KV-cache-size-matched experiments. The reason to do this is to give decode-focused methods such as MLA and GQA a fair chance. When we cannot perfectly match, we deliberately give the advantage for the non-CCA method. For instance, when comparing to MHA, we match parameters but ignore CCA's KV-cache compression rate. For MLA, GQA, we match parameters and compression rate, but not FLOPs since GQA and especially MLA utilize many more FLOPs than CCA.

In all ablations, the models were trained on a randomly selected subset of the Zyda2 \citep{zyda2} dataset. We trained standard Llama3-style 1B dense transformer models for 300B tokens and a 300M-active/1.5B-total parameter models using our MoE architecture. We tested both dense and MoE to determine if there were differences in trends between architectures. 

For CCGQA on our MoE architecture, we do a single experiment with $8\times$ compression to illustrate the Pareto frontier of increasing high arithmetic intensity with respect to performance (see fig \ref{fig:Architecture Comparison in the MoE Setting}). For different economic constraints one might opt for less arithmetic intensity, but a more computationally efficient variant.

\section{Results}

\begin{table*}[htbp]
\centering
\begin{tabular}{|ccc|ccccccc|c|}
\hline
\multicolumn{3}{|c|}{\textbf{Settings}} & \multicolumn{7}{c|}{\textbf{Evaluation Scores}} & \multicolumn{1}{c|}{\multirow{2}{*}{\textbf{Loss}}} \\
\cline{1-10}
\textbf{Convs} & \textbf{QK-Mean} & \textbf{V-Shift} & \textbf{HellaSwag} & \textbf{ARC Easy} & \textbf{ARC HARD} & \textbf{Piqa} & \textbf{Winogrande} & \textbf{Avg} & & \\
\hline
0 & \xmark & \xmark & 56.8 & 59.7 & 34.0 & 73.9 & 56.0 & 56.1 & & 2.330 \\
1 & \xmark & \xmark & 57.1 & 58.8 & 33.5 & 72.9 & 54.8 & 55.4 & & 2.327 \\
2 & \xmark & \xmark & 58.2 & 60.4 & 33.9 & 74.3 & 56.3 & 56.6 & & 2.319 \\
2 & \xmark & \cmark & 58.0 & 59.3 & 33.4 & 73.7 & 56.1 & 56.1 & & 2.317 \\
2 & \cmark & \cmark & 57.4 & 62.4 & 34.7 & 74.0 & 56.7 & 57.0 & & 2.315 \\
\hline
\end{tabular}
\caption{Ablation study of CCA in 1B parameter models. We test CCA with zero, one, or two conv layers, with and without qk-mean residual, and with and without the v-shift operation. Each version is trained on $\sim 300$ billion tokens of Zyda2 dataset, on which we show the validation cross-entropy loss and several benchmarks.}
\label{tab:table_3}
\end{table*}

\subsection{Dense Model Experiments}
For CCA on dense models, we show results for a configuration with 4 query heads and 4 kv heads. For CCGQA, we use 16 query heads and 4 kv heads, repeated 4 times. We also illustrate the compute tradeoff in the parameter-matched setting through a comparison between MLA, MHA, and GQA with CCA and CCGQA. Furthermore, for MLA, GQA, CCA and CCGQA, the KV-cache compression was matched at 1/4 of MHA. We observe that for dense models, in the parameter-matched setting, CCA outperforms MLA in loss while achieving a significant reduction in training and inference FLOPs. CCGQA outperforms MHA despite matching in parameters and with substantial KV-cache compression and reduced FLOPs. 
    \begin{figure}[htbp]
        \centering
        \includegraphics[width=0.5\textwidth]{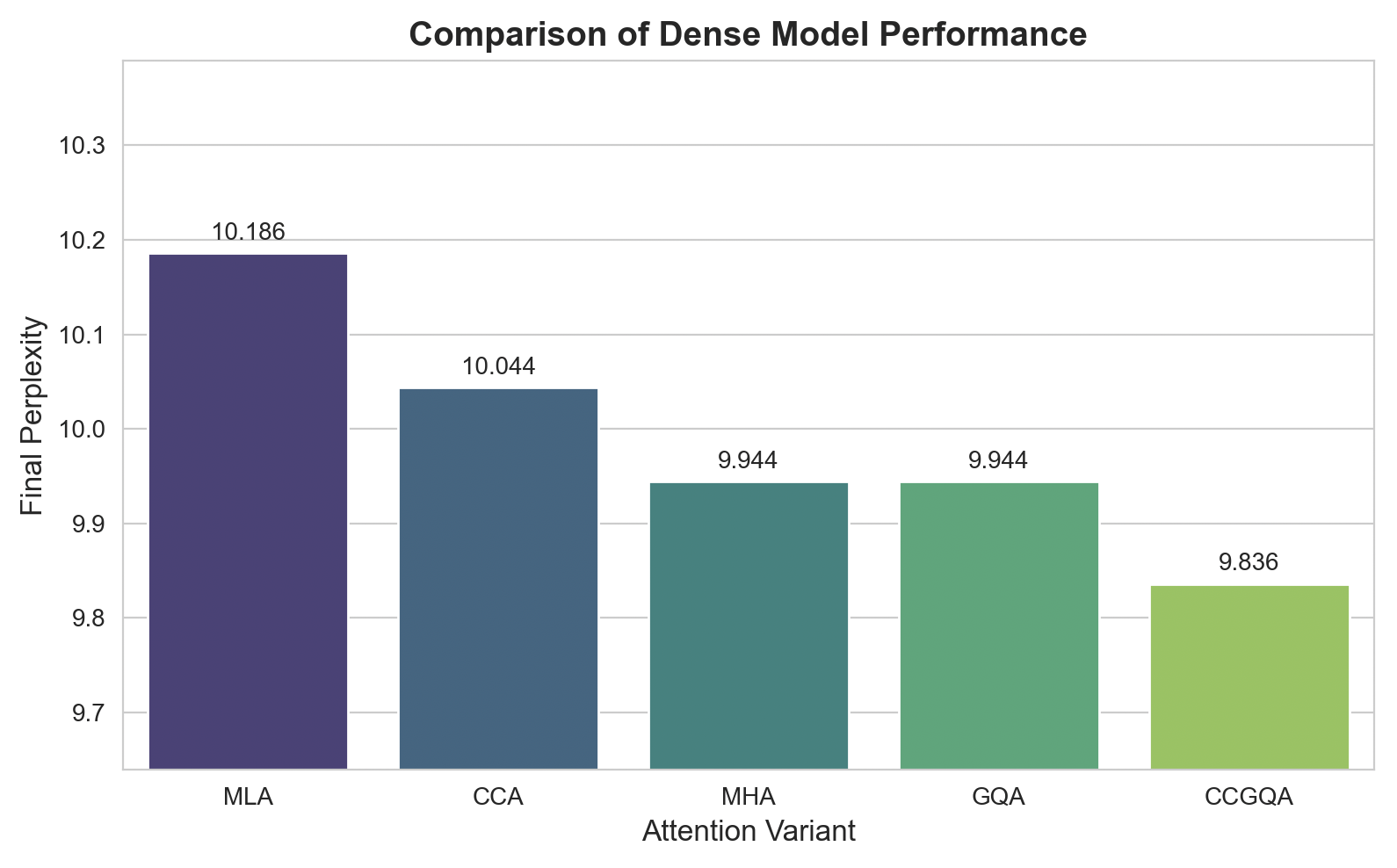}
        \caption{Comparison of perplexity on the Zyda2 dataset for 1B parameter dense transformer models trained for 300B tokens with different attention mechanisms. CCA beats MLA in the parameter-matched setting with less FLOPs. When matching CCA FLOPs to GQA and MHA via CCGQA, we see a substantial improvement in perplexity.}
        \label{fig: Architecture Comparison in the Dense Setting}
    \end{figure}

\subsection{MoE Model Experiments}
For CCA on our MoE architecture, we use CCGQA with 2 kv heads ($8\times$ compression) and 8 query heads ($2\times$ compression) and CCA with 4 q, kv heads ($4\times$ compression). These comparisons are all parameter-matched and KV-cache-matched with the exception of MHA, which has $4\times$ the KV-cache. We provide these results to further illustrate that parameter-efficient attention types, including MLA, are more useful in the MoE setting, since when parameter matching the forward and total parameters, reducing the parameters allocated to attention allows more to be assigned to the experts. This enables each expert to grow in size, which in turn reduces the total number of experts required for a fixed total parameter budget. Under our MoE architecture, we have found this reallocation to be beneficial. Additionally, shifting parameters from attention to the experts reduces the number of fixed parameters in the forward pass, making the model more expressive without increasing the total or forward parameter count.

For MLA, we match total parameters and forward parameters for training. In order to show that CCA's operations are universal across attention methods we additionally provide results for CCMLA in Appendix C. CCMLA utilizes the same query and KV-cache compression ratios as CCGQA, and crucially does not share key and value heads, as sequence convolutions on values prior to attention was empirically poor. We instead simply apply up-projections and 50\% rope to CCGQA to illustrate the tradeoff between additional expressivity in the up-projections versus decreased expressivity in the shared key rope across heads.

We show that CCA is capable of beating MLA with significantly less parameters and less FLOPs.
We show here that when matching KV-cache and decreasing compute with CCA, we beat MLA, and we can match MLA when halving KV-cache and matching compute with CCGQA, in the parameter-matched setting for MoEs. Depending on training versus inference cost, there exists a smooth Pareto frontier for which a CCGQA configuration is optimal.
    \begin{figure}[htbp]
        \centering        \includegraphics[width=0.5\textwidth]{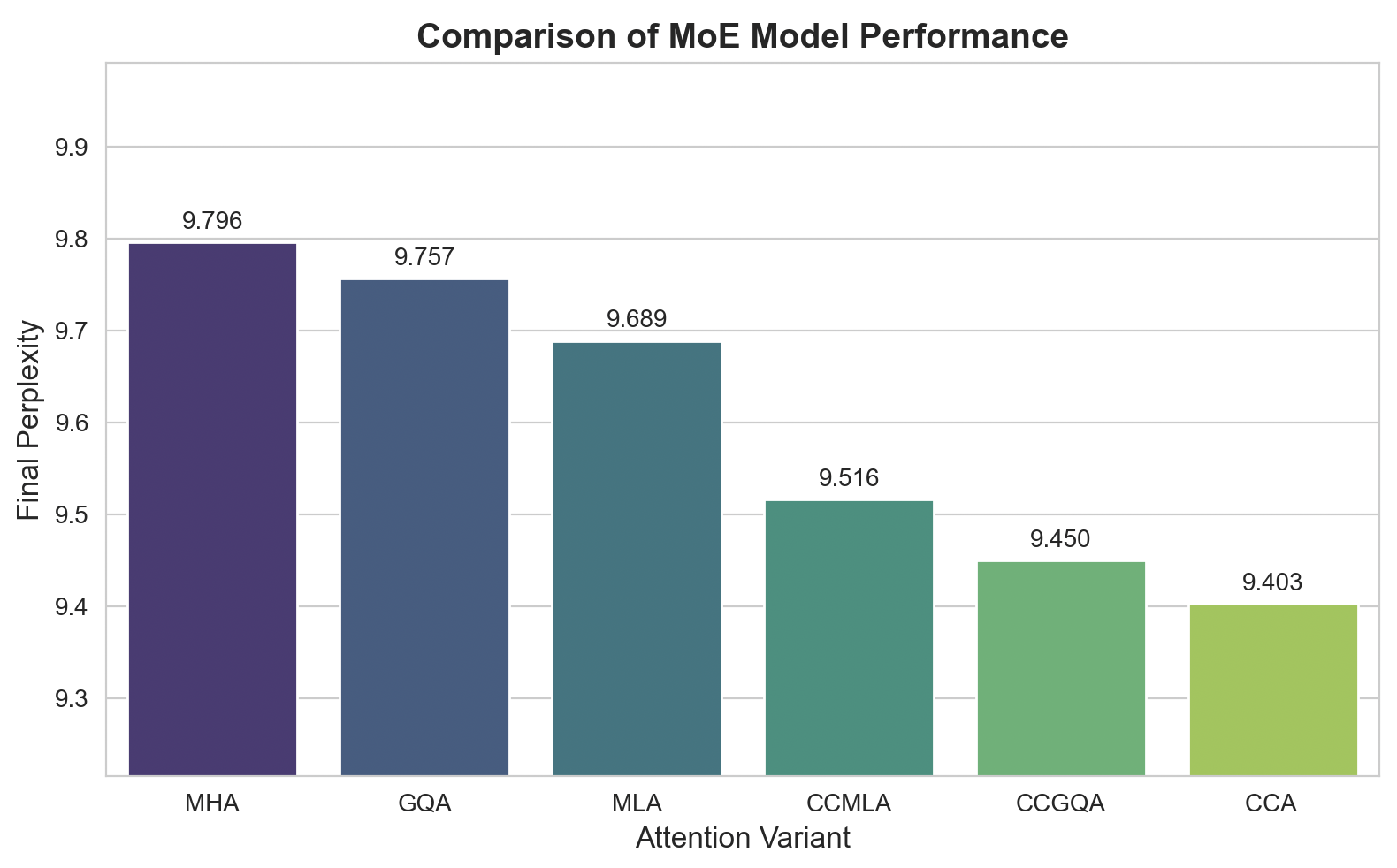}
        \caption{Comparison of perplexity on 50B tokens of the Zyda2 dataset for 350M/1.5B parameter proprietary MoE models with different attention mechanisms. Our proposed methods, CCA and CCGQA, achieve lower loss than GQA and MLA at equivalent parameter counts with less compute cost, and less training parameters in the case of MLA.}
        \label{fig:Architecture Comparison in the MoE Setting}
    \end{figure}

\subsection{CCA Ablations}

To illustrate the relative importance of each component in CCA, we show ablations on the 4 query, 4 key, 4 value variant of CCA on a 1B dense model and our proprietary 350M/1.5B ablation sized MoE. The majority of performance increase in CCA comes from the sequential convolutions. The auxiliary modifications of qk-mean and value-shift together provide a small, but noticeable decrease in perplexity and boost in evaluations. In our MoE, we see more significant jumps due to the auxiliary modifications, which can be illustrated here by perplexity shown in Table 4. 

\section{Kernel Performance}
\label{sec:kernel-and-system-performance}

\begin{figure*}
    \centering
    \includegraphics[width=.8\linewidth]{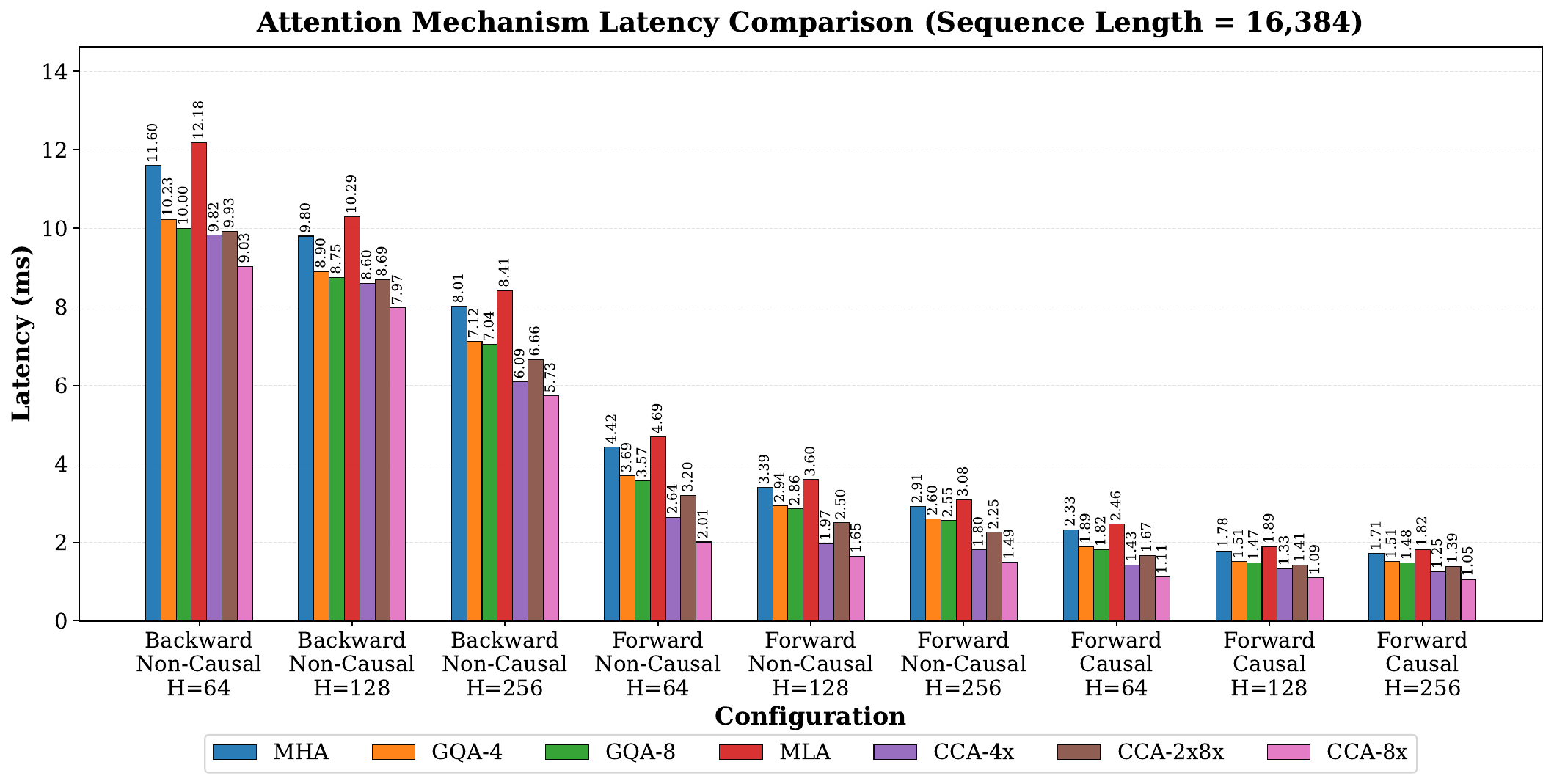}
    \caption{Performance of CCA versus competing attention methods with hidden dimension 2048 and BFLOAT16 on an H100 GPU}
    \label{fig:kernel}
\end{figure*}

Compressed Convolutional Attention (CCA) executes the entire attention operation in a compressed latent space, which significantly reduces both the arithmetic intensity and the data-movement requirements relative to competing methods such as MHA, GQA, and MLA. From \Cref{tab:attention_comparison}, we expect the $S^2$ terms from $QK^\top$ and $\mathrm{Attn}\!\cdot\!V$ to scale by $1/C$ once $Q$, $K$, and $V$ live in a latent of width $\tilde e=E/C$. We have designed an H100 GPU kernel to fuse the convolution with an online softmax in the style of the flash attention series of kernels \citep{dao2023flashattention2}.

\begin{figure*}[!htbp]
  \centering
  \includegraphics[width=0.6\textwidth]{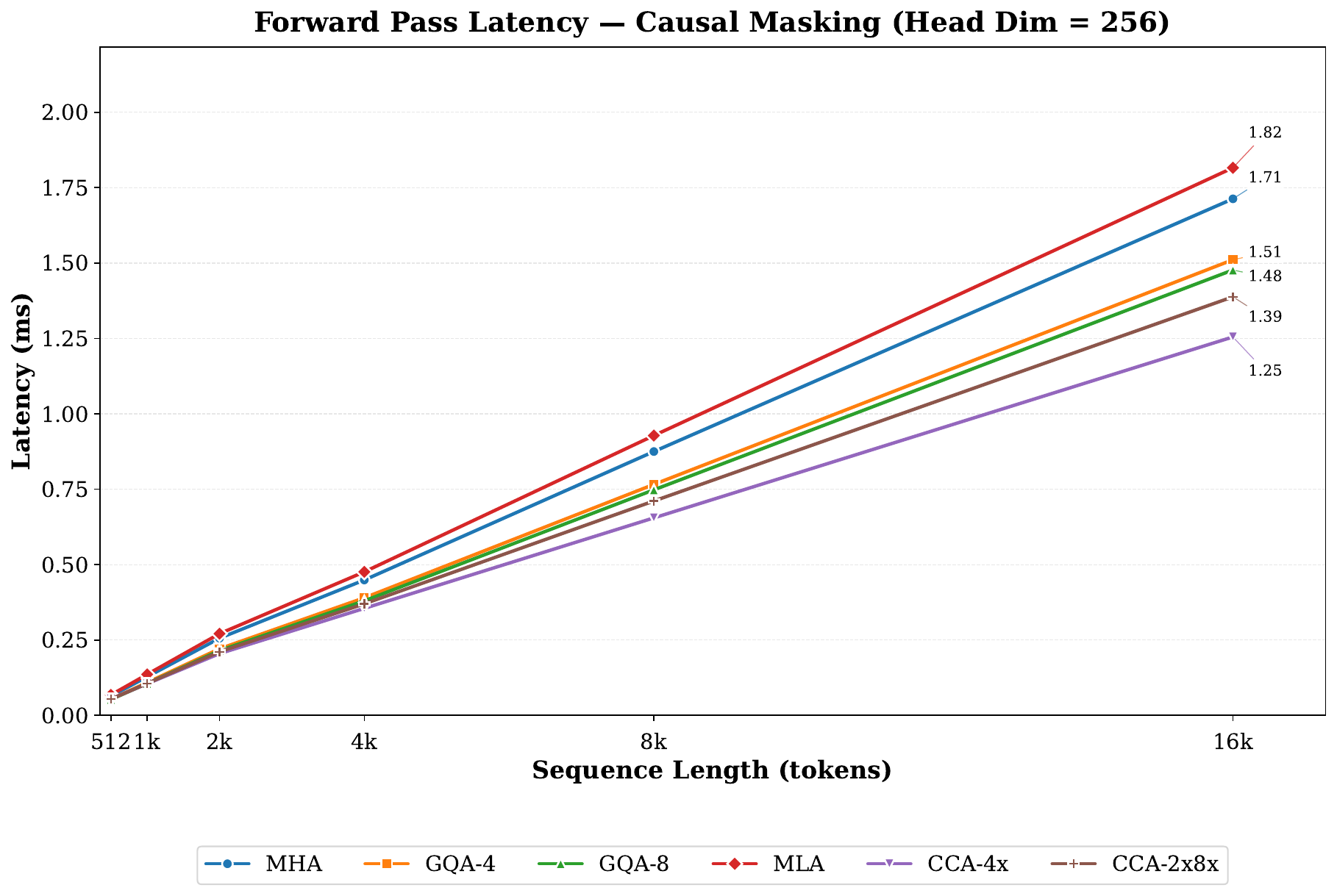}
  \caption{Prefill forward (causal) attention latencies: Head dimension 256}
  \label{fig:fwd-causal-256}
\end{figure*}

Our FLOP complexity model roughly aligns with the implementation results. Discounting the small convolutional terms called out in \Cref{tab:attention_comparison}, the $S^2$ components of both matrix multiplications shrink by $1/C$, and the projection terms shrink proportionally as well. As a result, we expect latency gains that grow with sequence length, since fixed overheads amortize at the larger sequence lengths.

A key difference from methods that only rebalance bandwidth in decode is that CCA reduces both prefill and decode compute and shrinks the KV cache simultaneously. CCGQA inherits GQA’s KV reuse benefits in the latent while preserving CCA’s $1/C$ scaling of the matmuls. KV cache results will be included in a follow-up work, since the focus of this initial work is on pretraining.

\begin{table}[htbp]
\centering
\begin{tabular}{|ccc|c|}
\hline
\multicolumn{3}{|c|}{\textbf{Ablation Settings}} & \multirow{2}{*}{\textbf{Loss}} \\
\cline{1-3}
\textbf{Convs} & \textbf{QK-Mean} & \textbf{V-Shift} & \\
\hline
0 & \xmark & \xmark & 2.280 \\
1 & \xmark & \xmark & 2.264 \\
2 & \xmark & \xmark & 2.252 \\
2 & \xmark & \cmark & 2.248 \\
2 & \cmark & \cmark & 2.241 \\
\hline
\end{tabular}
\caption{Ablation study of CCA in our 350M/1.5B MoE models. We test CCA with zero, one, or two conv layers, with and without qk-mean residual, and with and without the v-shift operation. Each version is trained on ~50 billion tokens of Zyda2 dataset, on which we show the validation cross-entropy loss.}
\label{tab:table_4}
\end{table}
\section{Discussion}

In this paper, we have presented CCA and CCGQA, a novel method for performing parameter compression in self-attention. CCA and CCGQA enable significant compression factors to be achieved while also improving performance relative to MHA, making significantly longer context lengths more feasible. CCA also outperforms existing methods such as GQA and MLA in model quality for fixed compression rate, allowing either greater compression rates to be used with acceptable loss or models to be improved at the same compression rate. Specifically, we find that in dense models we significantly outperform all other prominent attention types when matching training parameters to MHA and GQA (CCGQA, 16 query heads, 4 kv heads), and we beat MLA with $16\times$ less decoding FLOPs (CCA, 4 query heads, 4 kv heads), and with $8\times$ less FLOPs in the case of CCGQA. 

Our work is also the first to note and theoretically clarify that parameter-sharing methods such as GQA and parameter-compression methods such as MLA and CCA are orthogonal to one another and can be effectively combined. This implies that there is a Pareto-front which allows one to trade-off the degree of parameter sharing vs parameter compression to achieve the same compression rate.
Notably, CCA is the first method that performs attention \emph{entirely within a compressed latent space}, for both training and inference, thus demonstrating that there exists significant redundancy of parameters and activations in classic attention methods. This approach allows us to elegantly and seamlessly integrate RoPE (or indeed any position embedding) into CCA, unlike MLA, and also lets us save substantial compute for both prefill and decoding. Although CCA does not address the fundamental quadratic complexity of attention, by reducing the compute required by a factor of the compression rate, CCA can enable substantially longer sequences to be processed for a given compute and memory budget. 

Since it does not address the quadratic all-to-all nature of attention, CCA is thus orthogonal to methods that try to apply sequence compression like NSA \citep{yuan2025native}, MoBA \citep{lu2025mobamixtureblockattention}, and DSA \citep{deepseekai2024deepseekv32} and future work can explore to what extent we can apply both channel/cache compression and sequence compression and selectivity to continue to reduce the cost of attention while maintaining core sequence mixing performance. Future work could also investigate how well the compressed KV-cache of CCA interacts with offline KV-cache compression methods.

The architectural innovations used in CCA to boost performance -- sequence and channel mixing convolutions, qk-mean-adaptation, and value-shift, are not necessarily specific to CCA and could be used to improve sequence mixers in general. The value-shift adds a strong inductive bias that half the heads cannot see the present, which seems generally useful. RWKV \citep{peng2023rwkv} use a similar approach (which they call token-shift); which they find improves performance of their RNN-based sequence mixer. This provides some circumstantial evidence towards the general utility of this bias. The qk-mean effectively adds a bias towards strengthening the diagonal of the attention scores matrix and also, when combined with normalization, slightly sparsifies the key and query vectors. 

More generally, we expect that performing carefully chosen additional operations, including additional nonlinearity, on the compressed latent query, key, and value vectors is a promising direction to further increase the expressivity of attention and make up for the loss of expressivity from the dimensional compression. At high compression rates, such operations are relatively cheap in terms of FLOPs and parameters and, indeed in CCA, the convolutions and other operations have negligible overhead, since the vast majority of parameters exist in the downprojection matrices. However, despite being theoretically negligible in FLOPs, in practice these operations do often cause overhead in a naive pytorch implementation. A fused kernel for CCA is required to obtain the full speedup that is theoretically possible. One note for concern, however, is that compared to MLA/GQA, which like MHA, only do linear projections of q,k,v, CCA does more involved operations on the compressed latents. This induces a measure of inductive bias into CCA which is less present in MLA and GQA and may mean that CCA outperforms at small scales while the benefits lessen at larger scales. Testing how CCA performs and compares with alternatives on larger scales would be important to verify or falsify this hypothesis.

Due to CCGQA's versatility, it is also more amenable to support existing parallelism schemes, unlike MLA. For instance, sharding CCA's latent representation with TP incurs only the same cost as GQA and is relatively cheap as long as the TP rank is the same as the number of groups. Context-parallelism is also simple to implement, since one can simply communicate the smaller latent width $E/C$ instead of the standard full width $E$ within a ring or tree. Unlike MLA, CCA applies RoPE directly in the latent and only needs a constant-size latent halo for the causal convolutions and one-token value-shift, so no extra collectives or re-materialized up-projections are required.

\clearpage

\bibliographystyle{tmlr}
\bibliography{main}

@article{vaswani2017attention,
  title={Attention is all you need},
  author={Vaswani, Ashish and Shazeer, Noam and Parmar, Niki and Uszkoreit, Jakob and Jones, Llion and Gomez, Aidan N and Kaiser, {\L}ukasz and Polosukhin, Illia},
  journal={Advances in neural information processing systems},
  volume={30},
  year={2017}
}

@article{waleffe2024empirical,
  title={An Empirical Study of Mamba-based Language Models},
  author={Waleffe, Roger and Byeon, Wonmin and Riach, Duncan and Norick, Brandon and Korthikanti, Vijay and Dao, Tri and Gu, Albert and Hatamizadeh, Ali and Singh, Sudhakar and Narayanan, Deepak and others},
  journal={arXiv preprint arXiv:2406.07887},
  year={2024}
}

@inproceedings{katharopoulos2020transformers,
  title={Transformers are rnns: Fast autoregressive transformers with linear attention},
  author={Katharopoulos, Angelos and Vyas, Apoorv and Pappas, Nikolaos and Fleuret, Fran{\c{c}}ois},
  booktitle={International conference on machine learning},
  pages={5156--5165},
  year={2020},
  organization={PMLR}
}

@article{gu2023mamba,
  title={Mamba: Linear-time sequence modeling with selective state spaces},
  author={Gu, Albert and Dao, Tri},
  journal={arXiv preprint arXiv:2312.00752},
  year={2023}
}

@article{touvron2023llama,
  title={Llama 2: Open foundation and fine-tuned chat models},
  author={Touvron, Hugo and Martin, Louis and Stone, Kevin and Albert, Peter and Almahairi, Amjad and Babaei, Yasmine and Bashlykov, Nikolay and Batra, Soumya and Bhargava, Prajjwal and Bhosale, Shruti and others},
  journal={arXiv preprint arXiv:2307.09288},
  year={2023}
}

@article{jiang2023mistral,
  title={Mistral 7B},
  author={Jiang, Albert Q and Sablayrolles, Alexandre and Mensch, Arthur and Bamford, Chris and Chaplot, Devendra Singh and Casas, Diego de las and Bressand, Florian and Lengyel, Gianna and Lample, Guillaume and Saulnier, Lucile and others},
  journal={arXiv preprint arXiv:2310.06825},
  year={2023}
}

@article{gpt3,
  author       = {Tom B. Brown and
                  Benjamin Mann and
                  Nick Ryder and
                  Melanie Subbiah and
                  Jared Kaplan and
                  Prafulla Dhariwal and
                  Arvind Neelakantan and
                  Pranav Shyam and
                  Girish Sastry and
                  Amanda Askell and
                  Sandhini Agarwal and
                  Ariel Herbert{-}Voss and
                  Gretchen Krueger and
                  Tom Henighan and
                  Rewon Child and
                  Aditya Ramesh and
                  Daniel M. Ziegler and
                  Jeffrey Wu and
                  Clemens Winter and
                  Christopher Hesse and
                  Mark Chen and
                  Eric Sigler and
                  Mateusz Litwin and
                  Scott Gray and
                  Benjamin Chess and
                  Jack Clark and
                  Christopher Berner and
                  Sam McCandlish and
                  Alec Radford and
                  Ilya Sutskever and
                  Dario Amodei},
  title        = {Language Models are Few-Shot Learners},
  journal      = {CoRR},
  volume       = {abs/2005.14165},
  year         = {2020},
  url          = {https://arxiv.org/abs/2005.14165},
  eprinttype    = {arXiv},
  eprint       = {2005.14165},
  bibsource    = {dblp computer science bibliography, https://dblp.org}
}

@misc{yang2024gla,
      title={Gated Linear Attention Transformers with Hardware-Efficient Training}, 
      author={Songlin Yang and Bailin Wang and Yikang Shen and Rameswar Panda and Yoon Kim},
      year={2024},
      eprint={2312.06635},
      archivePrefix={arXiv},
      primaryClass={cs.LG},
      url={https://arxiv.org/abs/2312.06635}, 
}

@article{abdin2024phi,
  title={Phi-3 Technical Report: A Highly Capable Language Model Locally on Your Phone},
  author={Abdin, Marah and Jacobs, Sam Ade and Awan, Ammar Ahmad and Aneja, Jyoti and Awadallah, Ahmed and Awadalla, Hany and Bach, Nguyen and Bahree, Amit and Bakhtiari, Arash and Behl, Harkirat and others},
  journal={arXiv preprint arXiv:2404.14219},
  year={2024}
}

@article{gu2021efficiently,
  title={Efficiently modeling long sequences with structured state spaces},
  author={Gu, Albert and Goel, Karan and R{\'e}, Christopher},
  journal={arXiv preprint arXiv:2111.00396},
  year={2021}
}

@misc{Bamba,
  title = {Bamba: Inference-Efficient Hybrid Mamba2 Model},
  author = {Chu, Linsong and Kumari, Divya and Dao, Tri and Gu, Albert and Ganti, Raghu and Agrawal, Dakshi and Srivatsa, Mudhakar and Wertheimer, Davis and Lim, Yu Chin Fabian and Viros, Antoni and Gonzalez, Nelson and HoangTrong, Tuan and Arviv, Ofir and Perlitz, Yotam and Shmueli, Michal and Shen, Haochen and Zhang, Minjia and Goodhart, Gabe and Wang, Naigang and Hill, Nick and Rosenkranz, Joshua and Liu, Chi-Chun and Hoque, Adnan and Yang, Chih-Chieh and Sharma, Sukriti and Uong, Anh and Gala, Jay and Zawad, Syed and Gordon, Ryan},
  year = {2024},
  month = {December},
  publisher = {HuggingFace},
  url = {https://huggingface.co/blog/bamba}
}

@misc{grazzi2024mamba,
      title={Is Mamba Capable of In-Context Learning?}, 
      author={Riccardo Grazzi and Julien Siems and Simon Schrodi and Thomas Brox and Frank Hutter},
      year={2024},
      eprint={2402.03170},
      archivePrefix={arXiv},
      primaryClass={cs.LG}
}

@misc{park2024mamba,
      title={Can Mamba Learn How to Learn? A Comparative Study on In-Context Learning Tasks}, 
      author={Jongho Park and Jaeseung Park and Zheyang Xiong and Nayoung Lee and Jaewoong Cho and Samet Oymak and Kangwook Lee and Dimitris Papailiopoulos},
      year={2024},
      eprint={2402.04248},
      archivePrefix={arXiv},
      primaryClass={cs.LG}
}

@misc{jelassi2024repeat,
      title={Repeat After Me: Transformers are Better than State Space Models at Copying}, 
      author={Samy Jelassi and David Brandfonbrener and Sham M. Kakade and Eran Malach},
      year={2024},
      eprint={2402.01032},
      archivePrefix={arXiv},
      primaryClass={cs.LG}
}

@misc{lieber2024jamba,
      title={Jamba: A Hybrid Transformer-Mamba Language Model}, 
      author={Opher Lieber and Barak Lenz and Hofit Bata and Gal Cohen and Jhonathan Osin and Itay Dalmedigos and Erez Safahi and Shaked Meirom and Yonatan Belinkov and Shai Shalev-Shwartz and Omri Abend and Raz Alon and Tomer Asida and Amir Bergman and Roman Glozman and Michael Gokhman and Avashalom Manevich and Nir Ratner and Noam Rozen and Erez Shwartz and Mor Zusman and Yoav Shoham},
      year={2024},
      eprint={2403.19887},
      archivePrefix={arXiv},
      primaryClass={cs.CL}
}

@misc{peng2023rwkv,
      title={RWKV: Reinventing RNNs for the Transformer Era}, 
      author={Bo Peng and Eric Alcaide and Quentin Anthony and Alon Albalak and Samuel Arcadinho and Stella Biderman and Huanqi Cao and Xin Cheng and Michael Chung and Matteo Grella and Kranthi Kiran GV and Xuzheng He and Haowen Hou and Jiaju Lin and Przemyslaw Kazienko and Jan Kocon and Jiaming Kong and Bartlomiej Koptyra and Hayden Lau and Krishna Sri Ipsit Mantri and Ferdinand Mom and Atsushi Saito and Guangyu Song and Xiangru Tang and Bolun Wang and Johan S. Wind and Stanislaw Wozniak and Ruichong Zhang and Zhenyuan Zhang and Qihang Zhao and Peng Zhou and Qinghua Zhou and Jian Zhu and Rui-Jie Zhu},
      year={2023},
      eprint={2305.13048},
      archivePrefix={arXiv},
      primaryClass={cs.CL}
}

@misc{sun2023retentive,
      title={Retentive Network: A Successor to Transformer for Large Language Models}, 
      author={Yutao Sun and Li Dong and Shaohan Huang and Shuming Ma and Yuqing Xia and Jilong Xue and Jianyong Wang and Furu Wei},
      year={2023},
      eprint={2307.08621},
      archivePrefix={arXiv},
      primaryClass={cs.CL}
}

@article{kaplan2020scaling,
  title={Scaling laws for neural language models},
  author={Kaplan, Jared and McCandlish, Sam and Henighan, Tom and Brown, Tom B and Chess, Benjamin and Child, Rewon and Gray, Scott and Radford, Alec and Wu, Jeffrey and Amodei, Dario},
  journal={arXiv preprint arXiv:2001.08361},
  year={2020},
  url={https://arxiv.org/abs/2001.08361}
}

@misc{dao2023flashattention2,
      title={FlashAttention-2: Faster Attention with Better Parallelism and Work Partitioning}, 
      author={Tri Dao},
      year={2023},
      eprint={2307.08691},
      archivePrefix={arXiv},
      primaryClass={cs.LG}
}

@misc{h100,
    title={ NVIDIA H100 tensor core gpu architecture}, 
    author={NVIDIA},
    year={2022},
    url={https://resources.nvidia.com/en-us-hopper-architecture/nvidia-h100-tensor-c}
}

@misc{glorioso2024zamba,
      title={{Zamba: A Compact 7B SSM Hybrid Model}}, 
      author={Paolo Glorioso and Quentin Anthony and Yury Tokpanov and James Whittington and Jonathan Pilault and Adam Ibrahim and Beren Millidge},
      year={2024},
      eprint={2405.16712},
      archivePrefix={arXiv},
      primaryClass={cs.LG},
      url={https://arxiv.org/abs/2405.16712}, 
}

@misc{su2023rotary,
      title={RoFormer: Enhanced Transformer with Rotary Position Embedding}, 
      author={Jianlin Su and Yu Lu and Shengfeng Pan and Ahmed Murtadha and Bo Wen and Yunfeng Liu},
      year={2023},
      eprint={2104.09864},
      archivePrefix={arXiv},
      primaryClass={cs.CL},
      url={https://arxiv.org/abs/2104.09864}, 
}

@misc{zyda2,
    author = {Yury Tokpanov, Paolo Glorioso and Ayush Dattagupta and Vibhu Jawa and Ryan Wolf and Vikranth Jeyakumar and Arham Mehta, Quentin Anthony and Beren Millidge},
    title = {Building {Zyda-2}, a 5 {Trillion} {Token} {High-Quality} {Dataset}, with {NVIDIA} {NeMo} {Curator}},
    url = {https://www.zyphra.com/post/building-zyda-2},
    publisher = {Zyphra},
    year = {2024},
    month = {October},
    day = {15}
}

@article{guo2025deepseek,
  title={Deepseek-r1: Incentivizing reasoning capability in llms via reinforcement learning},
  author={Guo, Daya and Yang, Dejian and Zhang, Haowei and Song, Junxiao and Zhang, Ruoyu and Xu, Runxin and Zhu, Qihao and Ma, Shirong and Wang, Peiyi and Bi, Xiao and others},
  journal={arXiv preprint arXiv:2501.12948},
  year={2025}
}

@article{team2025kimi,
  title={Kimi K2: Open Agentic Intelligence},
  author={Team, Kimi and Bai, Yifan and Bao, Yiping and Chen, Guanduo and Chen, Jiahao and Chen, Ningxin and Chen, Ruijue and Chen, Yanru and Chen, Yuankun and Chen, Yutian and others},
  journal={arXiv preprint arXiv:2507.20534},
  year={2025}
}

@article{liu2023ring,
  title={Ring attention with blockwise transformers for near-infinite context},
  author={Liu, Hao and Zaharia, Matei and Abbeel, Pieter},
  journal={arXiv preprint arXiv:2310.01889},
  year={2023}
}

@article{ge2023context,
  title={In-context autoencoder for context compression in a large language model},
  author={Ge, Tao and Hu, Jing and Wang, Lei and Wang, Xun and Chen, Si-Qing and Wei, Furu},
  journal={arXiv preprint arXiv:2307.06945},
  year={2023}
}

@article{yang2024lossless,
  title={Lossless KV Cache Compression to 2\%},
  author={Yang, Zhen and Han, JN and Wu, Kan and Xie, Ruobing and Wang, An and Sun, Xingwu and Kang, Zhanhui},
  journal={arXiv preprint arXiv:2410.15252},
  year={2024}
}

@article{Zhu2025canon,
  title={Physics of Language Models: Part 4.1, Architecture Design and the Magic of Canon Layers},
  author={Zeyuan Allen-Zhu},
  journal={SSRN Electronic Journal},
  year={2025},
  url={https://papers.ssrn.com/sol3/papers.cfm?abstract_id=5240330}
}

@article{kim2024lexico,
  title={Lexico: Extreme KV Cache Compression via Sparse Coding over Universal Dictionaries},
  author={Kim, Junhyuck and Park, Jongho and Cho, Jaewoong and Papailiopoulos, Dimitris},
  journal={arXiv preprint arXiv:2412.08890},
  year={2024}
}

@article{liu2024clusterkv,
  title={Clusterkv: Manipulating llm kv cache in semantic space for recallable compression},
  author={Liu, Guangda and Li, Chengwei and Zhao, Jieru and Zhang, Chenqi and Guo, Minyi},
  journal={arXiv preprint arXiv:2412.03213},
  year={2024}
}

@article{yuan2025native,
  title={Native sparse attention: Hardware-aligned and natively trainable sparse attention},
  author={Yuan, Jingyang and Gao, Huazuo and Dai, Damai and Luo, Junyu and Zhao, Liang and Zhang, Zhengyan and Xie, Zhenda and Wei, YX and Wang, Lean and Xiao, Zhiping and others},
  journal={arXiv preprint arXiv:2502.11089},
  year={2025}
}

@article{ainslie2023gqa,
  title={Gqa: Training generalized multi-query transformer models from multi-head checkpoints},
  author={Ainslie, Joshua and Lee-Thorp, James and De Jong, Michiel and Zemlyanskiy, Yury and Lebr{\'o}n, Federico and Sanghai, Sumit},
  journal={arXiv preprint arXiv:2305.13245},
  year={2023}
}

@article{shazeer2019fast,
  title={Fast transformer decoding: One write-head is all you need},
  author={Shazeer, Noam},
  journal={arXiv preprint arXiv:1911.02150},
  year={2019}
}

@article{shyam2024tree,
  title={Tree attention: Topology-aware decoding for long-context attention on gpu clusters},
  author={Shyam, Vasudev and Pilault, Jonathan and Shepperd, Emily and Anthony, Quentin and Millidge, Beren},
  journal={arXiv preprint arXiv:2408.04093},
  year={2024}
}

@article{lambert2024t,
  title={T$\backslash$" ulu 3: Pushing frontiers in open language model post-training},
  author={Lambert, Nathan and Morrison, Jacob and Pyatkin, Valentina and Huang, Shengyi and Ivison, Hamish and Brahman, Faeze and Miranda, Lester James V and Liu, Alisa and Dziri, Nouha and Lyu, Shane and others},
  journal={arXiv preprint arXiv:2411.15124},
  year={2024}
}

@article{glorioso2024zamba2,
  title={The Zamba2 Suite: Technical Report},
  author={Glorioso, Paolo and Anthony, Quentin and Tokpanov, Yury and Golubeva, Anna and Shyam, Vasudev and Whittington, James and Pilault, Jonathan and Millidge, Beren},
  journal={arXiv preprint arXiv:2411.15242},
  year={2024}
}

@misc{deepseekv3,
      title={DeepSeek-V3 Technical Report}, 
      author={DeepSeek-AI},
      year={2025},
      eprint={2412.19437},
      archivePrefix={arXiv},
      primaryClass={cs.CL},
      url={https://arxiv.org/abs/2412.19437}, 
}

@misc{lu2025mobamixtureblockattention,
      title={MoBA: Mixture of Block Attention for Long-Context LLMs}, 
      author={Enzhe Lu and Zhejun Jiang and Jingyuan Liu and Yulun Du and Tao Jiang and Chao Hong and Shaowei Liu and Weiran He and Enming Yuan and Yuzhi Wang and Zhiqi Huang and Huan Yuan and Suting Xu and Xinran Xu and Guokun Lai and Yanru Chen and Huabin Zheng and Junjie Yan and Jianlin Su and Yuxin Wu and Neo Y. Zhang and Zhilin Yang and Xinyu Zhou and Mingxing Zhang and Jiezhong Qiu},
      year={2025},
      eprint={2502.13189},
      archivePrefix={arXiv},
      primaryClass={cs.LG},
      url={https://arxiv.org/abs/2502.13189}, 
}

@inproceedings{wang2024lengthgeneralizationcausaltransformers,
author = {Wang, Jie and Ji, Tao and Wu, Yuanbin and Yan, Hang and Gui, Tao and Huang, Xuanjing and Wang, Xiaoling},
year = {2024},
month = {01},
pages = {14024-14040},
title = {Length Generalization of Causal Transformers without Position Encoding},
doi = {10.18653/v1/2024.findings-acl.834}
}

@misc{deepseekai2024deepseekv32,
      title={DeepSeek-V3.2-Exp: Boosting Long-Context Efficiency with DeepSeek Sparse Attention}, 
      author={DeepSeek-AI},
      year={2025},
}

\clearpage



\section*{A.1: PyTorch Code for CCA}

\begin{strip}
\begin{lstlisting}[
    language=Python,
    caption={PyTorch implementation of Compressed Convolutional Attention (CCA).},
    label={code:cca},
    basicstyle=\ttfamily\small,
    keywordstyle=\color{blue}\bfseries,
    commentstyle=\color{green!50!black},
    stringstyle=\color{orange},
    showstringspaces=false,
    numbers=left
]
class CCA(nn.Module):
    def forward(self, hidden_states):
        # Note: for simplicity this code is split up; many operations can be fused.
        # hidden_states.shape: seq_len, batch_size, hidden_dim

        # Initial low-rank combined down projection for query and key
        qk_packed0 = self.linear_qk(hidden_states)[0]
        qk_packed1 = qk_packed0.permute(1, 2, 0)

        # Padding for causal convolutions
        qk_packed2 = F.pad(qk_packed1, (self.total_padding, 0))

        # Sequence convolution
        qk_packed3 = self.conv_qk0(qk_packed2)

        # Head-wise grouped sequence convolution
        qk_packed3 = self.conv_qk1(qk_packed3).permute(2, 0, 1)

        # Calculation of query and key mean for biasing
        key_pre = qk_packed0[..., self.latent_q_dim:(self.latent_k_dim + self.latent_q_dim)]
        key_pre = key_pre.view(*key_pre.shape[:2], self.num_k_heads, self.head_dim).unsqueeze(-2).repeat(1, 1, 1, self.gqa_groups, 1)
        key_pre = key_pre.view(*key_pre.shape[:2], self.num_q_heads, self.head_dim)
        query_pre = qk_packed0[..., 0:self.latent_q_dim]
        query_pre = query_pre.view(*query_pre.shape[:2], self.num_q_heads, self.head_dim)
        qk_mean_q = (query_pre + key_pre) / 2
        qk_mean_k = qk_mean_q.view(*qk_mean_q.shape[:2], self.num_k_heads, self.gqa_groups, -1).mean(dim=-2)
        query = qk_packed3[..., 0:self.latent_q_dim].reshape(* qk_packed3.shape[:2], self.num_q_heads, -1) + qk_mean_q
        key = qk_packed3[..., self.latent_q_dim:(self.latent_k_dim + self.latent_q_dim)].reshape(* qk_packed3.shape[:2], self.num_k_heads, -1) + qk_mean_k

        # Padding for value_shift
        hidden_states_d = F.pad(hidden_states[:-1], pad=(0, 0, 0, 0, 1, 0))

        # Value_shift
        value1, _ = self.val_proj1(hidden_states)
        value2, _ = self.val_proj2(hidden_states_d)
        value = torch.cat([value1, value2], dim=-1).reshape(* hidden_states.shape[:2], self.num_k_heads, -1)

        # Query and key normalization with temperature and scaling
        query_norm = query.norm(p=2, dim=-1, keepdim=True)
        key_norm = key.norm(p=2, dim=-1, keepdim=True)
        key = (key * self.sqrt_head_dim / key_norm) * torch.exp(self.temp[None, None].unsqueeze(-1))
        query = (query * self.sqrt_head_dim / query_norm)

        # Can now pass to flash attention with requisite settings
        return query.contiguous(), key.contiguous(), value.contiguous()
\end{lstlisting}
\end{strip}

\section*{B.1: MLA Inference Compute Considerations}
\label{sec:appendix-mla-inference}

\begin{figure*}[t!]
        \centering
        \includegraphics[width=.7\textwidth]{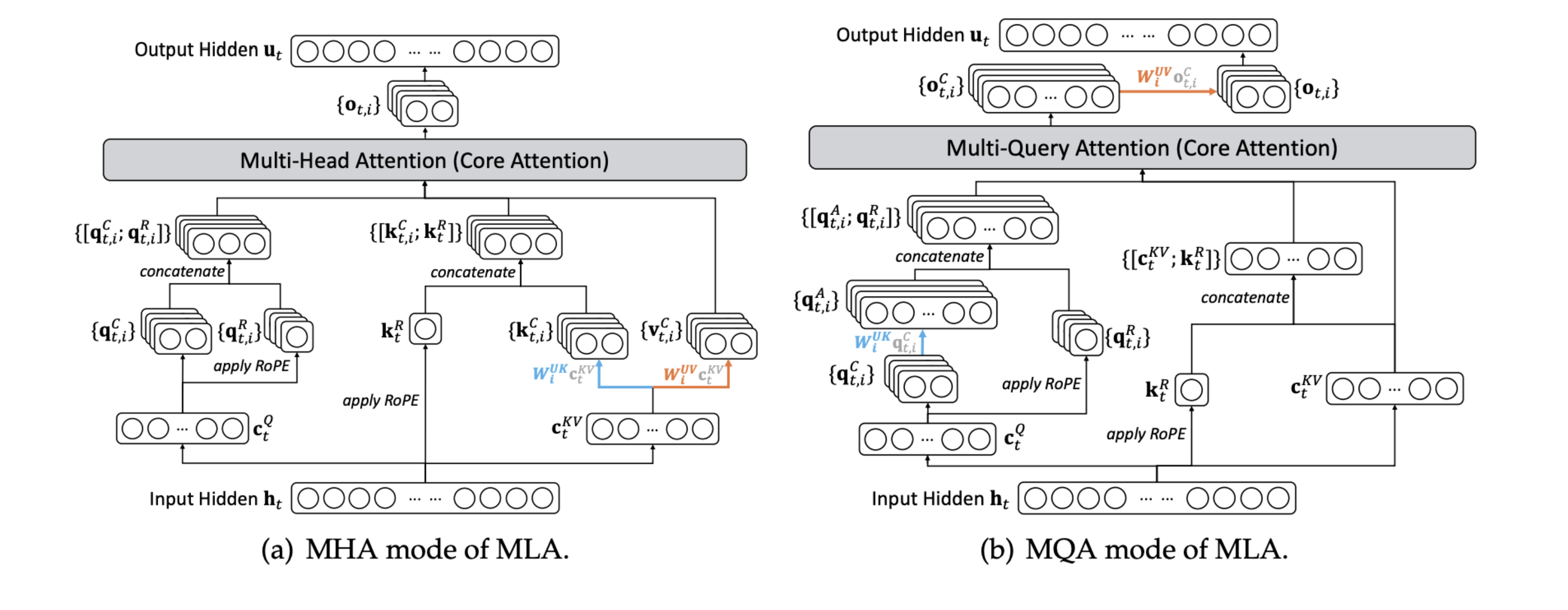}
        \caption{Depiction of MLA in both MHA and MQA modes from \textbf{\citet{deepseekai2024deepseekv32}}}
        \label{fig:MLA-MQA-MHA}
\end{figure*}

MLA inference is somewhat underdiscussed in the literature, but it is a prominent attention used in modern frontier models \citep{deepseekv3,team2025kimi}.

The logic behind MLA is as follows: for decoding, the optimal attention architecture is MQA due to being memory bandwidth bound, and for training/prefill the optimal attention architecture is MHA so as to maximize expressivity. The two modes of MLA enable switching between these settings as desired -- in the decode setting, the MQA variant can be used while in prefill, or training settings, the standard MHA-style MLA can be used. The full MHA-style MLA is slightly less expensive than regular MHA in terms of both FLOPs and parameters, due to low rank projections, but MLA still retains the traditional self-attention compute cost. 

Another way to think about this is that MQA maximizes the utilization of memory bandwidth by performing repeated computation over the shared key-value heads, while MHA maximizes unique computations for expressivity but at the cost of massive memory bandwidth requirements. MLA has a shared key-value MQA in the decoding mode that leverages high arithmetic intensity proportional to the number of heads. This falls short in cases where speculative decoding is utilized such that the arithmetic intesity passes the roofline; MLA additionally fails in cases where tensor parallel (TP) is used in inference, as MQA is no longer advantageous due to the necessity of repeating the shared KV-cache across devices. The optimal inference attention with tensor parallel is a latent GQA with number of key-value heads being equivalent to TP size, while having as many groups as possible to maximize arithmetic intensity. CCGQA is thus extremely well-suited for tensor-parallelism compared with MLA.

MLA has an arithmetic intensity of $2n_{heads}$ at inference time, which is very large, and targets the ridge of the roofline plot on a H100 for a single query. For speculative decoding, the excess use of FLOPs for relatively little increase in performance is not necessarily optimal. The arithmetic intensity required to breach the ridge of the roofline for an Nvidia H100 under bfloat16 is ~295 FLOPs per byte \citep{h100}. Deepseek seemingly chose their number of heads for the DeepseekV3 model to accordingly saturate the roofline to approach compute bound inference at batch size 1. GQA has a limited arithmetic intensity of $n_{groups}$, but is far more versatile under a number of conditions. 
Recall MLA at inference time:
\begin{align}
\mathrm{softmax}(\underbrace{q_{lr}^\top W^q W_k^\top}_{q' \in \mathbb{R}^{n_h \times d_c}}\underbrace{kv}_{k'^\top \in \mathbb{R}^{d_c\times N}}) \underbrace{kv^\top}_{v' \in \mathbb{R}^{N \times d_c}} W^v W^o\
\end{align}

This is equivalent to MQA with the head size being the shared KV latent. This is optimal when setting number of heads exactly to the ridge of the roofline plot. In most cases, using MQA mode is a large increase in FLOP count, but these FLOPS are computed by streaming multiprocessors (SMs) that would otherwise go unused or idle on decoding. With an approximate arithmetic intensity of 2$n_{heads}$, Deepseek selects their number of heads to almost exactly reach the ridge of the roofline plot. Considering Deepseek are under constraints of H800 devices, using MQA mode makes sense, since tensor parallel is heavily penalized. Specifically, under tensor parallel (TP) conditions. E.g. under TP=8, with an MQA model with 1 kv head and 16 query heads, the model must copy the shared kv head across devices according to the TP rank. The optimal attention has as many unique kv heads as TP shards, as to maximize kv reuse within devices and minimize kv reuse across devices. During the creation of CCA and CCGQA, this was heavily considered. This also appears to be why models which use MLA tend towards heavy expert-parallelism and ultimately pipeline parallelism instead of tensor parallelism.

An additional point is that while MLA is capable of better compute utilization on decode, the increase in FLOPs that would otherwise go unused does not automatically result in victory. Model quality and latency, not SM utilization, is the end goal. The primary tradeoff to mode-switch between MQA and MHA in traditional GQA attentions is whether maximizing of unique computations in training mode (MHA) is worth the decrease in expressivity from shared key rope and sharing keys and values. In order to show that this decreased expressivity imposes a steep penalty in practice, we test a CCMLA variant with up-projections on the latent key and value, and with shared key rope, as shown in figure \ref{fig:Architecture Comparison in the MoE Setting}. These results illustrate that it may not be necessarily optimal to utilize MLA's architecture under many conditions and that CCGQA is both more flexible and more expressive.

We designed CCA to be versatile and agnostic to all future model parallelism strategies and speculative decoding strategies, with arithmetic intensity of $n_{groups}$. Since CCA is capable of much larger KV-cache compression with the same performance, this enables a larger arithmetic intensity compared to GQA if desired. 

\begin{figure*}[htbp]
        \centering
        \includegraphics[width=.6\textwidth]{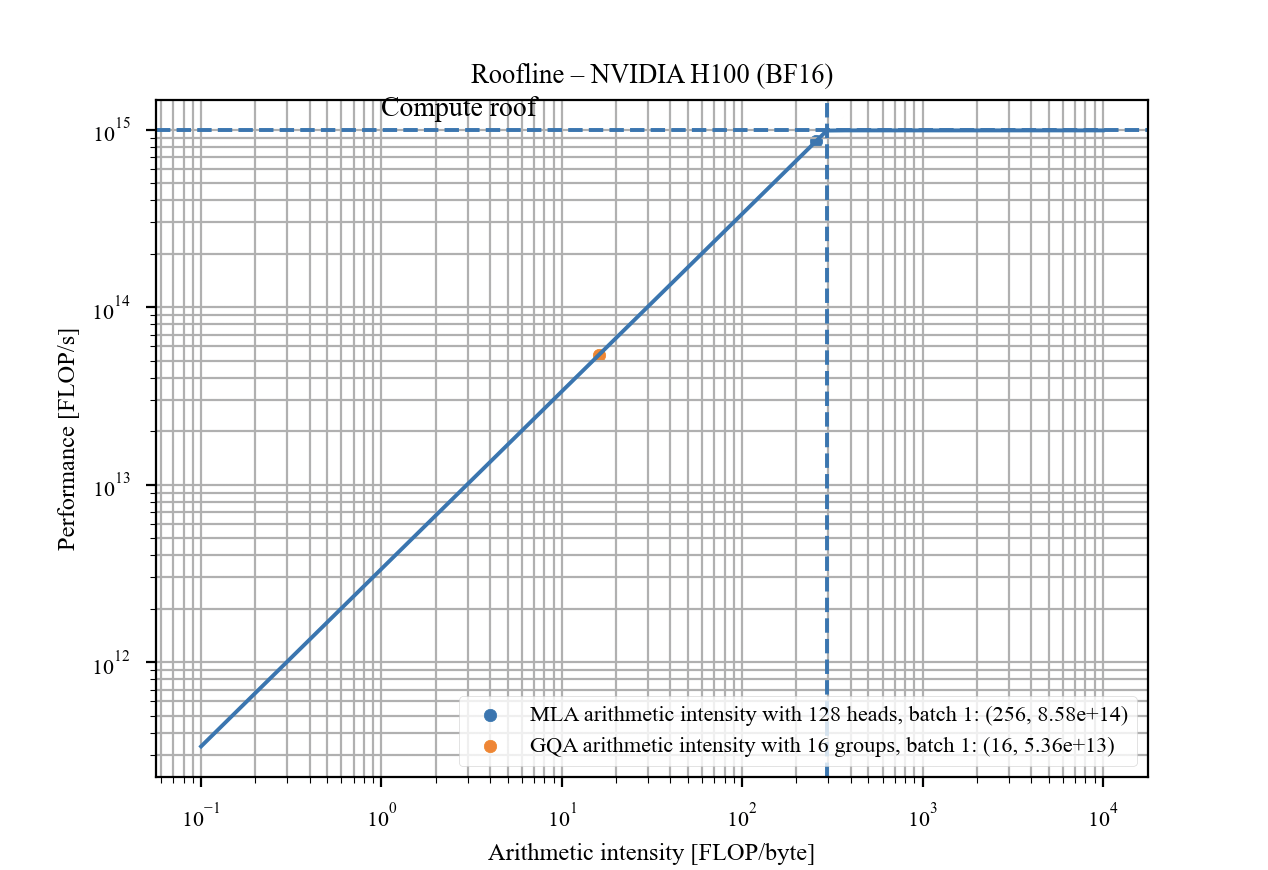}
        \caption{Dense BF16 roofline of an Nvidia Hopper H100 SXM GPU, with the theoretical arithmetic intensity for both GQA with 16 groups, and MLA with 128 heads. Note CCGQA has the same arithmetic intensity as GQA.}
        \label{fig:roofline}
    \end{figure*}


\section*{C.1 Loss Plots}
Here, we present our full loss plots for CCA under both MoE and dense conditions.
For all of our MoE ablations we use a 350M active/1.5B total parameter model with 28 layers. Our 1B parameter dense model was trained with 24 layers for 300B tokens, while our small MoE was trained for 50B tokens. Both models were trained on a subset of the Zyda2 dataset. We parameter match all comparisons across both total parameters and active parameters for inference. This ensures a fair comparison between attentions with drastically different higher arithmetic intensity.

\begin{figure*}[htbp]
  \centering
  \begin{subfigure}[htbp]{0.48\textwidth}
    \centering
    \includegraphics[width=\linewidth]{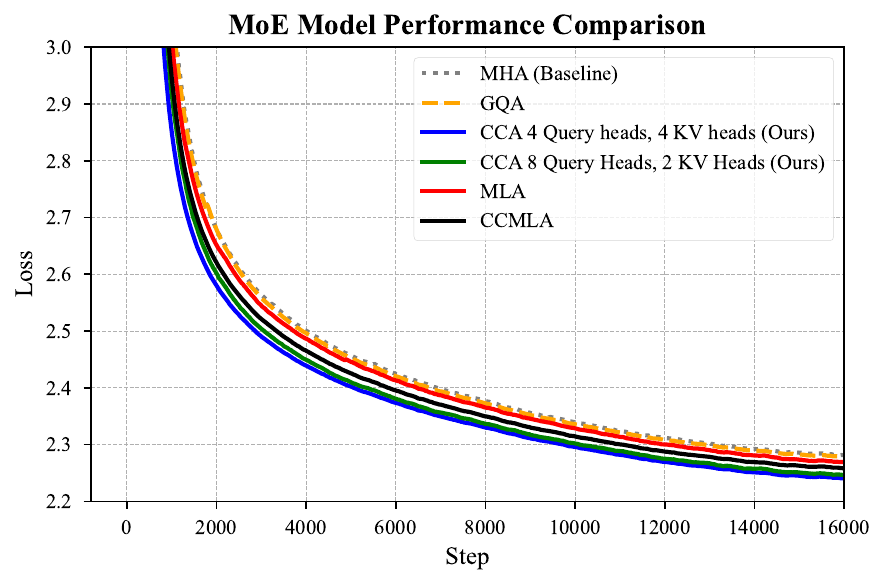}
    \caption{MoE Model Performance - Un-zoomed view}
    \label{fig:moe_results_nozoom}
  \end{subfigure}
  \hspace{0.02\textwidth}
  \begin{subfigure}[htbp]{0.48\textwidth}
    \centering
    \includegraphics[width=\linewidth]{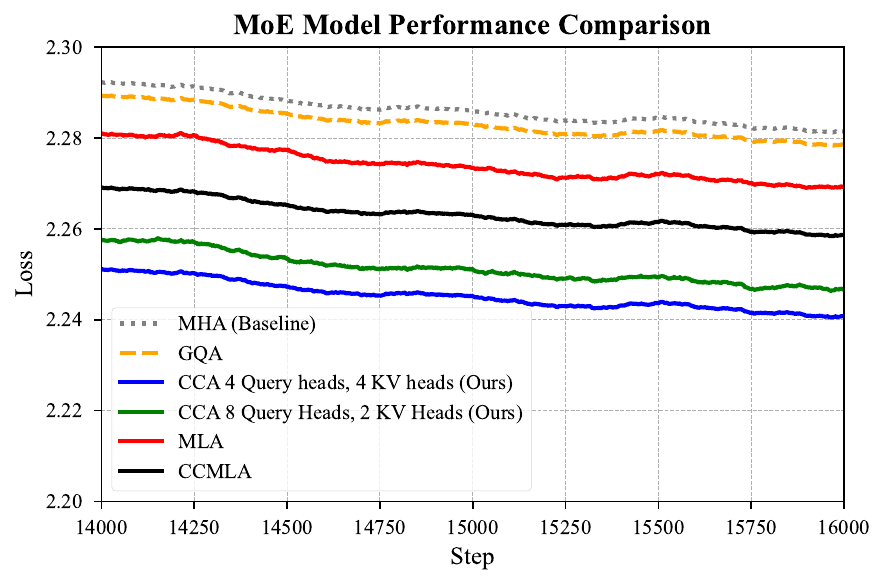}
    \caption{MoE Model Performance - Zoomed view}
    \label{fig:moe_results_zoom}
  \end{subfigure}
  
  \vspace{15pt} 
  
  \begin{subfigure}[htbp]{0.48\textwidth}
    \centering
    \includegraphics[width=\linewidth]{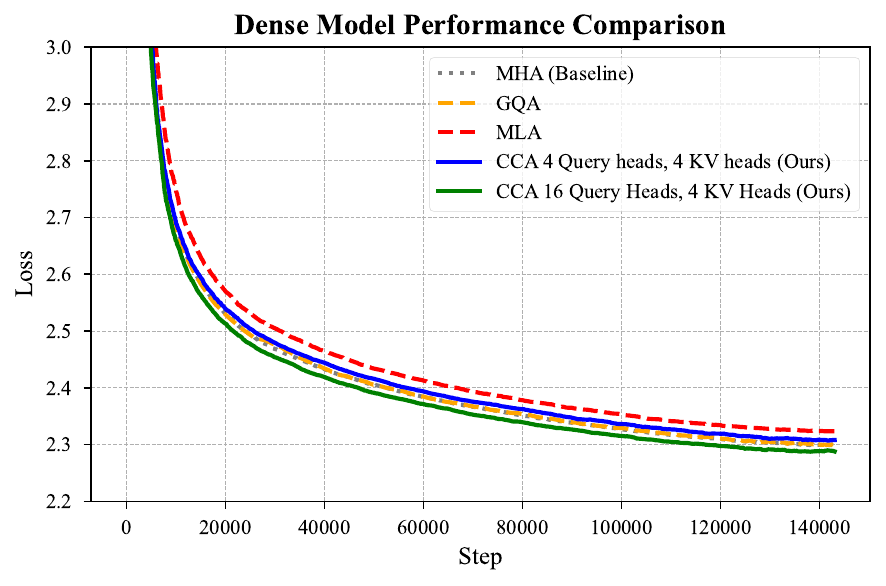}
    \caption{Dense Model Performance - Un-zoomed view}
    \label{fig:densenozoom}
  \end{subfigure}
  \hspace{0.02\textwidth}
  \begin{subfigure}[htbp]{0.48\textwidth}
    \centering
    \includegraphics[width=\linewidth]{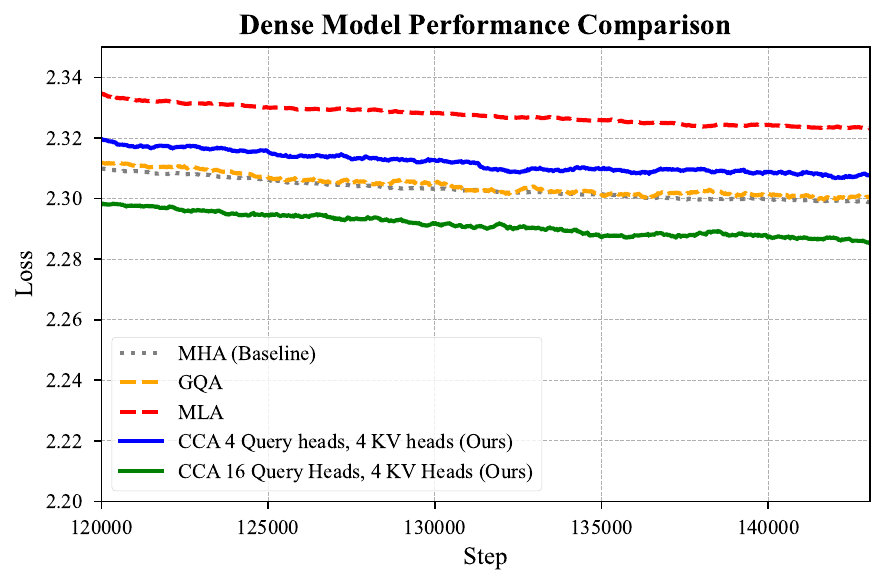}
    \caption{Dense Model Performance - Zoomed view}
    \label{fig:dense_ablation_zoom}
  \end{subfigure}
  
  \caption{Comparison of loss on the Zyda2 dataset for both MoE (top row) and dense (bottom row) models. For MoE: 350M/1.5B parameter models with different attention mechanisms. Our proposed methods, CCA and CCGQA, achieve lower loss than existing efficient attention variants like GQA and MLA at equivalent parameter counts. For Dense: 1B parameter models where all attention methods are restricted to equal KV-cache size (4x compression) except MHA. CCGQA has 8x KV-cache compression, while MLA, CCA and GQA have 4x compression relative to MHA.}
  \label{fig:all_results}
\end{figure*}

Notice that training was stable in all cases and that the differences in loss appear stable across many steps. Once they emerge they do not appear to converge again or cross one-another, meaning that some methods have long-term advantages over others in loss, averaged across samples. Per sample, the losses track each other very closely but at a roughly fixed offset.

\clearpage
\FloatBarrier

\begingroup
\raggedbottom

\section*{D.1 Benchmarking Setup and Kernel Notes}

All measurements are single-GPU latencies on an H100 with BF16 precision, using $E{=}2048$, head\_dim $\in \{64,128,256\}$, and sequence lengths from $512$ to $16{,}384$. The latency of the forward is reported in both non-causal and causal forms; backward results are reported for non-causal. CCA and CCGQA execute attention entirely in the latent space of width $\tilde e=E/C$, using RoPE position embeddings, L2 normalization for the queries and keys with key temperature scaling, qk-mean coupling, and value-shift as fused prologues/epilogues with online softmax for the causal case.

The alignment between the theoretical FLOP predictions and latencies is clearest at large $S$, where the $S^2/C$ terms dominate. At shorter sequences, constant overheads from kernel launches, reductions, and the fixed parts of prologue/epilogue keep the realized speedups below the ideal factor $C$, but the ordering across methods remains stable. MLA’s decode-time amortization strategy and GQA’s KV sharing primarily reconfigure bandwidth usage; neither reduces the core $S^2$ arithmetic during prefill or training, which is why their curves sit closer to MHA than CCA in the forward plots.

While most prior works comparing attention implementations use TFLOPs, we chose latency. This is because, while TFLOPs are useful to compare an implementation's achieved efficiency compared to the maximum throughput of the accelerator for a given operation, it is misleading to compare TFLOPs between different attention methods entirely, since the actual FLOP costs of the methods are different. CCA is expected to have slightly lower throughput on large accelerators than MHA doe to the reduced compute needs, but this reduction in compute itself leads to significant end-to-end speedup and better performance in terms of loss. In general, efficiently using a lot of flops is only useful if the FLOPs buy you something. \looseness=-1  Using substantially less FLOPs, slightly less efficiently, can often be useful in practice. We will also note that another benefit of performing attention on a smaller latent space, reduction in KV-cache size, is not demonstrated in these results but will result in significant decoding speed improvements.

\endgroup


\begin{figure*}[!htbp]
  \centering
  \includegraphics[width=0.7\textwidth]{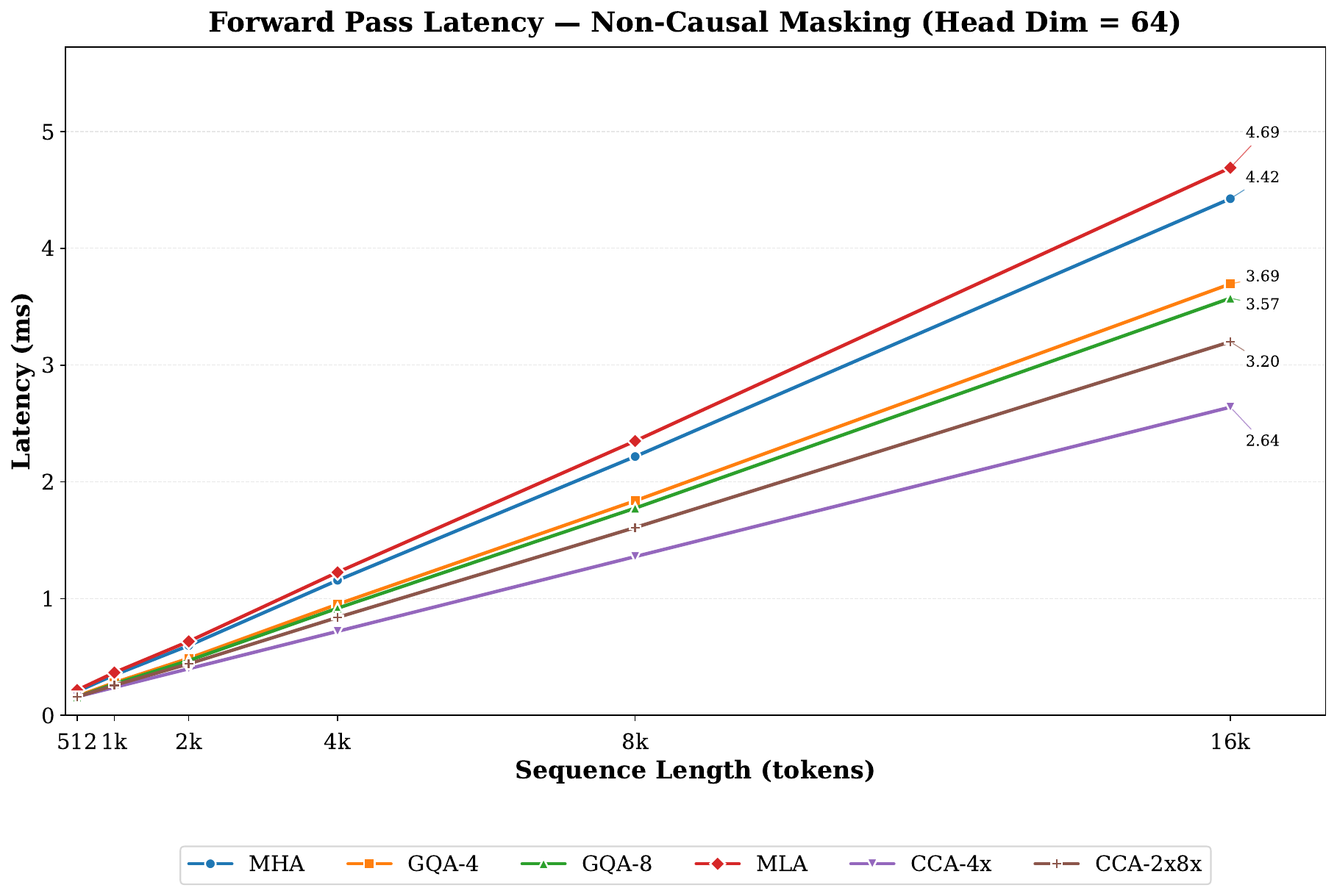}
  \caption{Prefill forward (non-causal) attention latencies: Head dimension 64}
  \label{fig:fwd-nocausal-64}
\end{figure*}

\begin{figure*}[!htbp]
  \centering
  \includegraphics[width=0.7\textwidth]{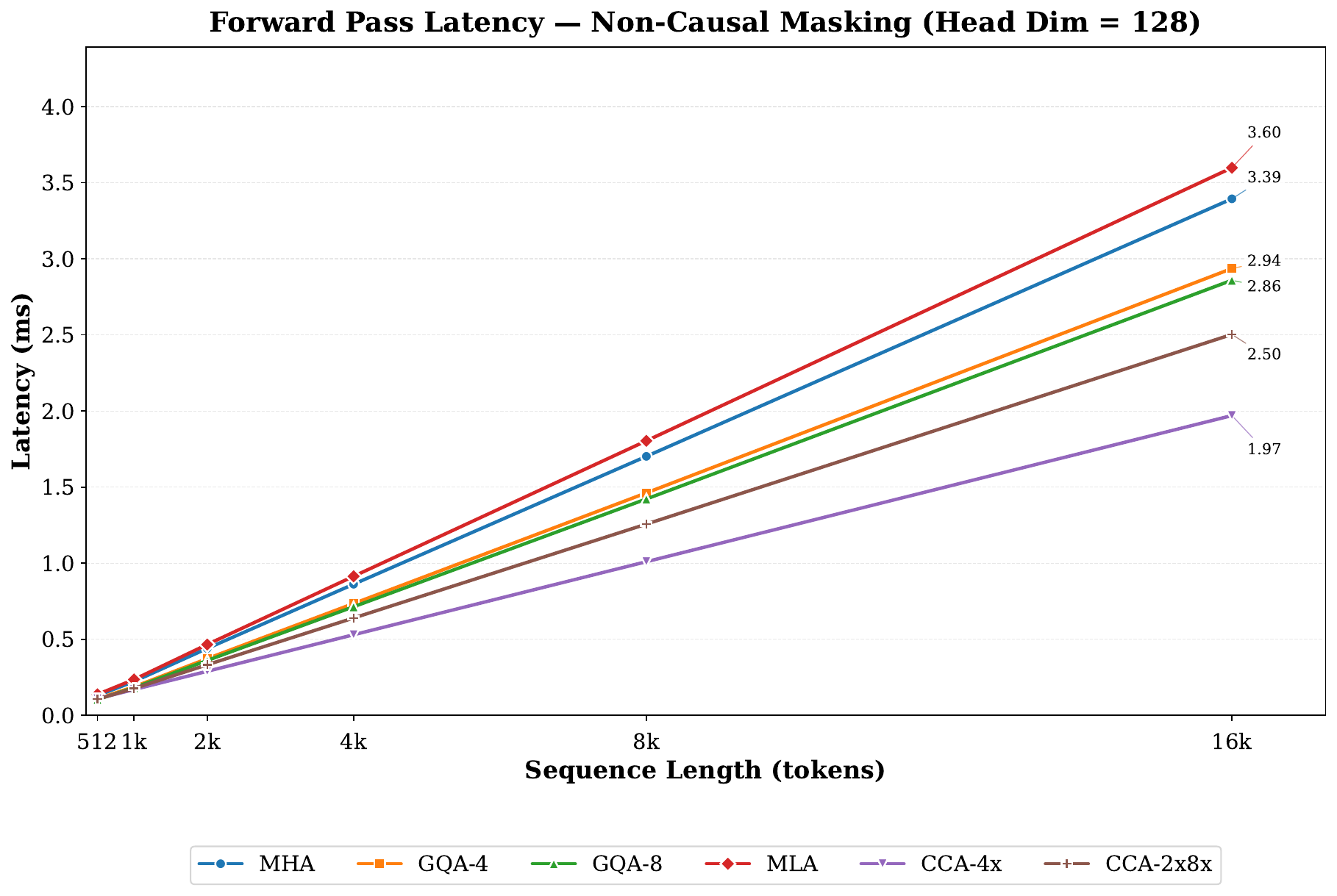}
  \caption{Prefill forward (non-causal) attention latencies: Head dimension 128}
  \label{fig:fwd-nocausal-128}
\end{figure*}

\begin{figure*}[!htbp]
  \centering
  \includegraphics[width=0.7\textwidth]{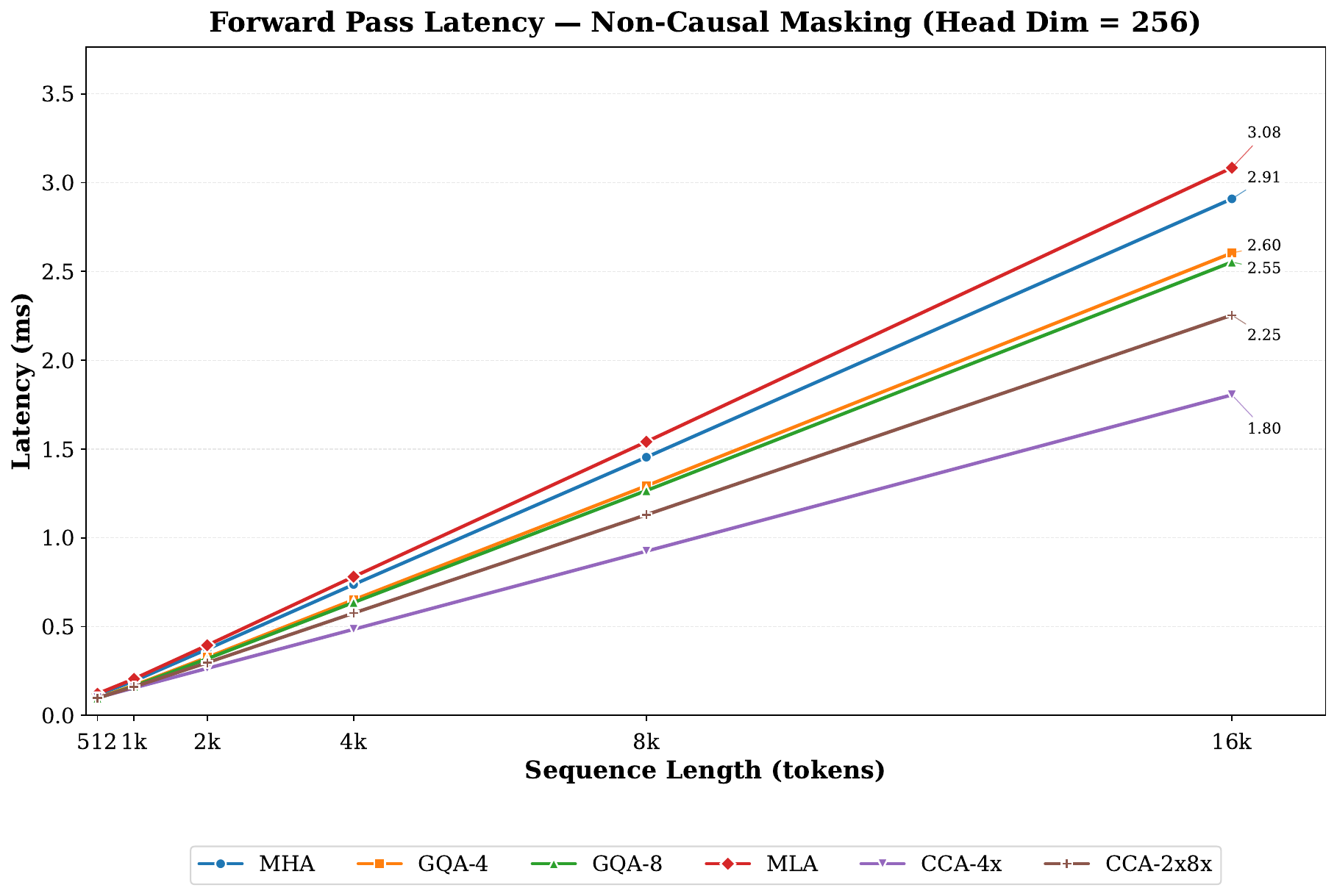}
  \caption{Prefill forward (non-causal) attention latencies: Head dimension 256}
  \label{fig:fwd-nocausal-256}
\end{figure*}

\begin{figure*}[!htbp]
  \centering
  \includegraphics[width=0.7\textwidth]{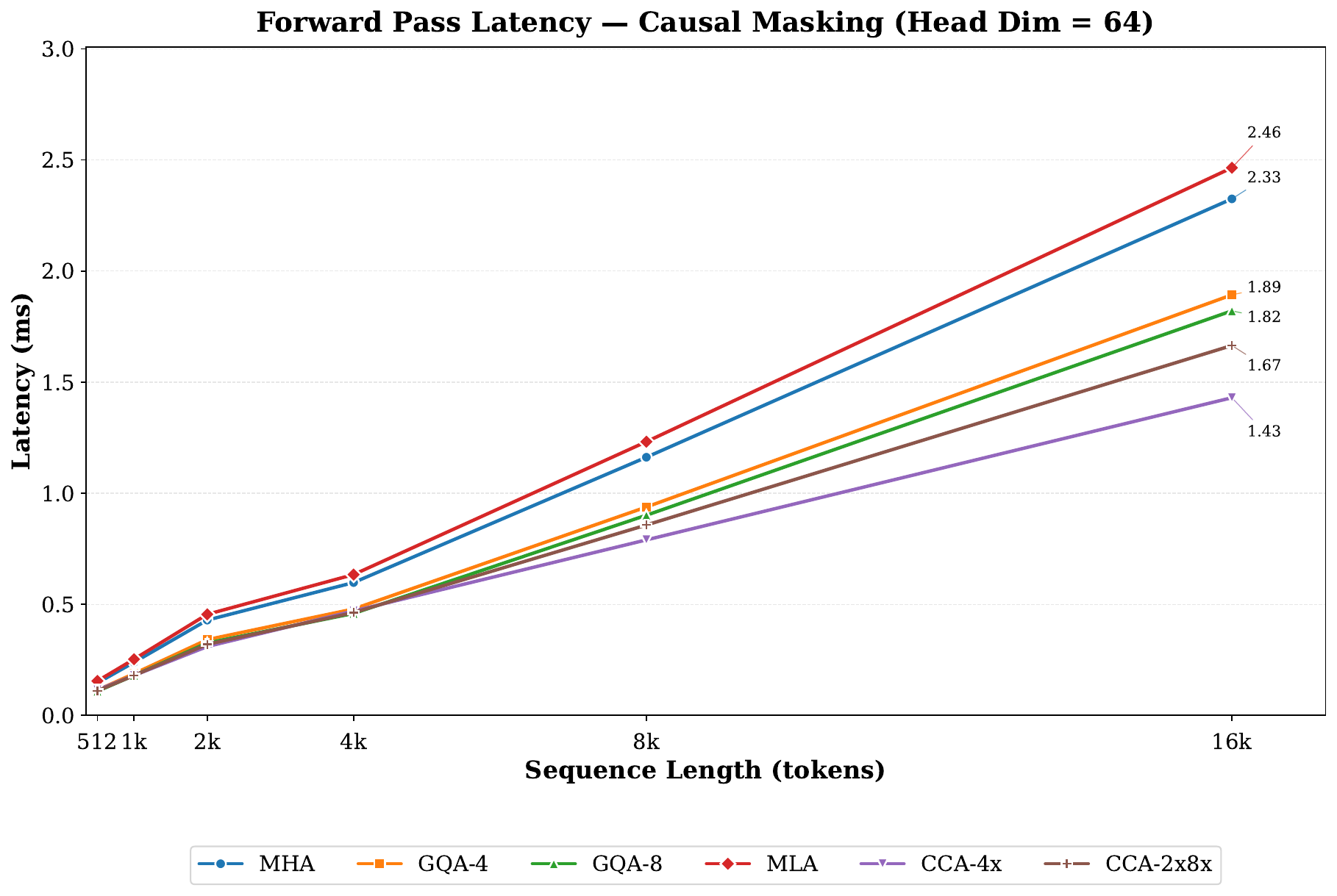}
  \caption{Prefill forward (causal) attention latencies: Head dimension 64}
  \label{fig:fwd-causal-64}
\end{figure*}

\begin{figure*}[!htbp]
  \centering
  \includegraphics[width=0.7\textwidth]{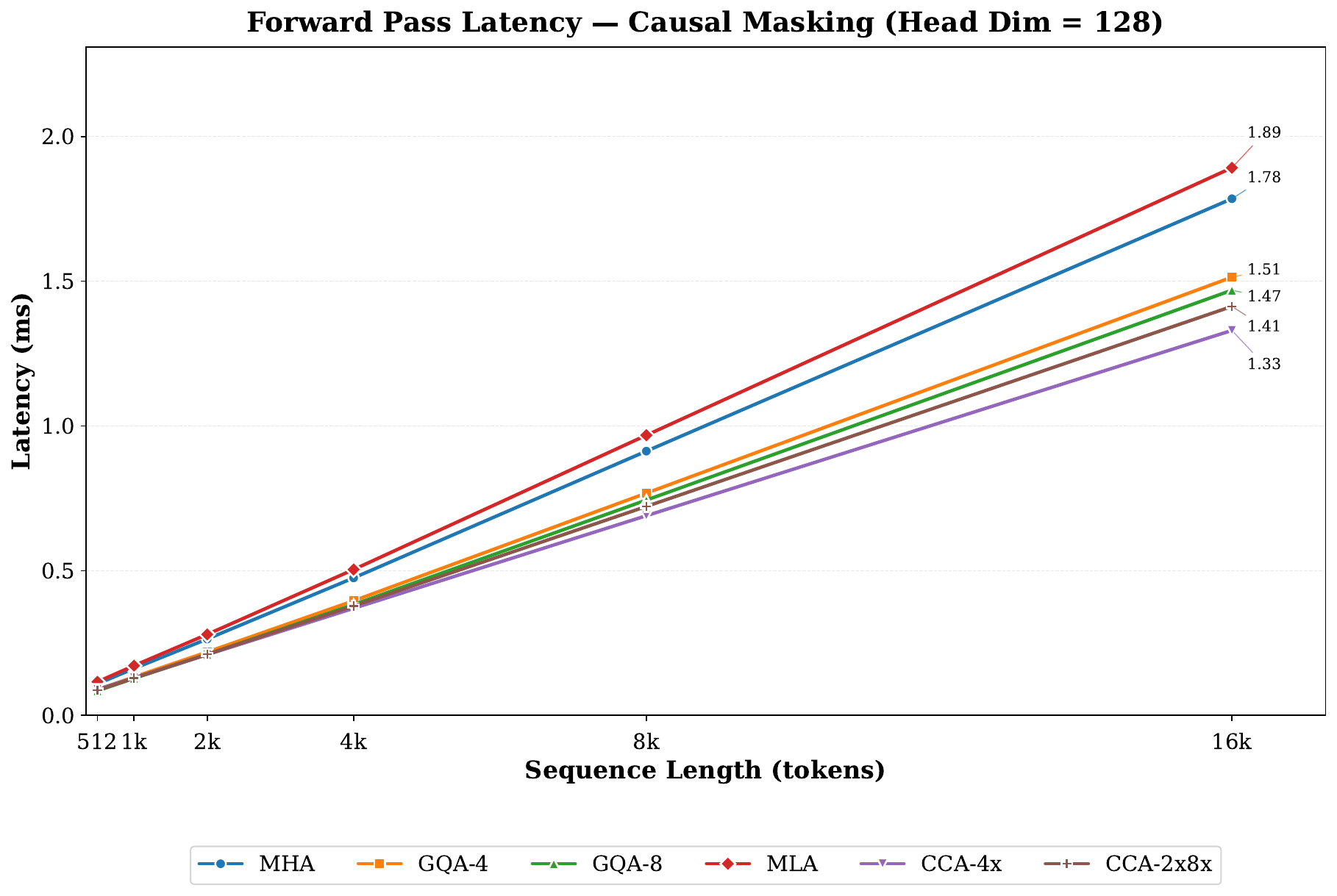}
  \caption{Prefill forward (causal) attention latencies: Head dimension 128}
  \label{fig:fwd-causal-128}
\end{figure*}

\begin{figure*}[!htbp]
  \centering
  \includegraphics[width=0.7\textwidth]{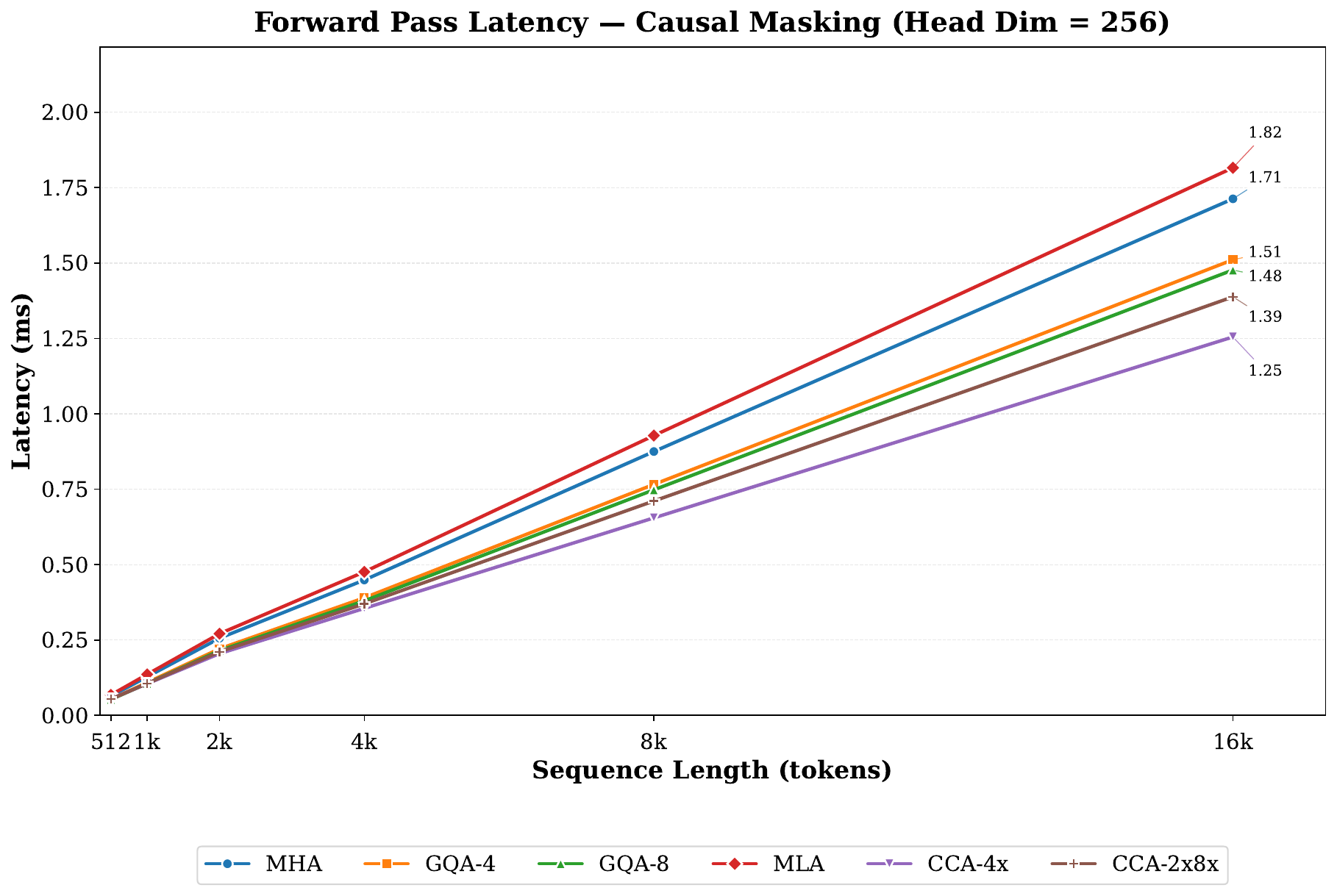}
  \caption{Prefill forward (causal) attention latencies: Head dimension 256}
  \label{fig:fwd-causal-256}
\end{figure*}

\begin{figure*}[!htbp]
  \centering
  \includegraphics[width=0.7\textwidth]{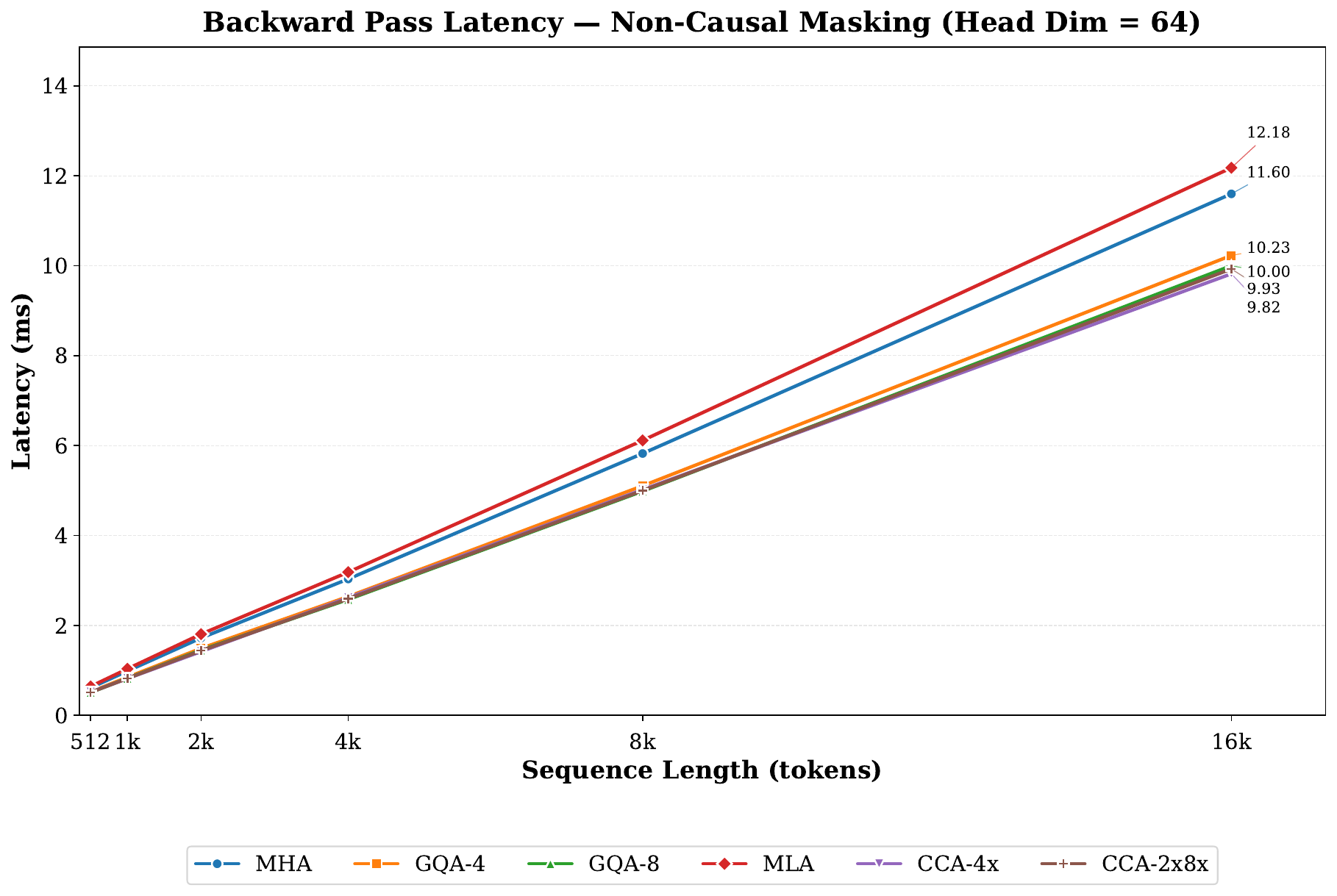}
  \caption{Backward attention latencies: Head dimension 64}
  \label{fig:bwd-64}
\end{figure*}

\begin{figure*}[!htbp]
  \centering
  \includegraphics[width=0.7\textwidth]{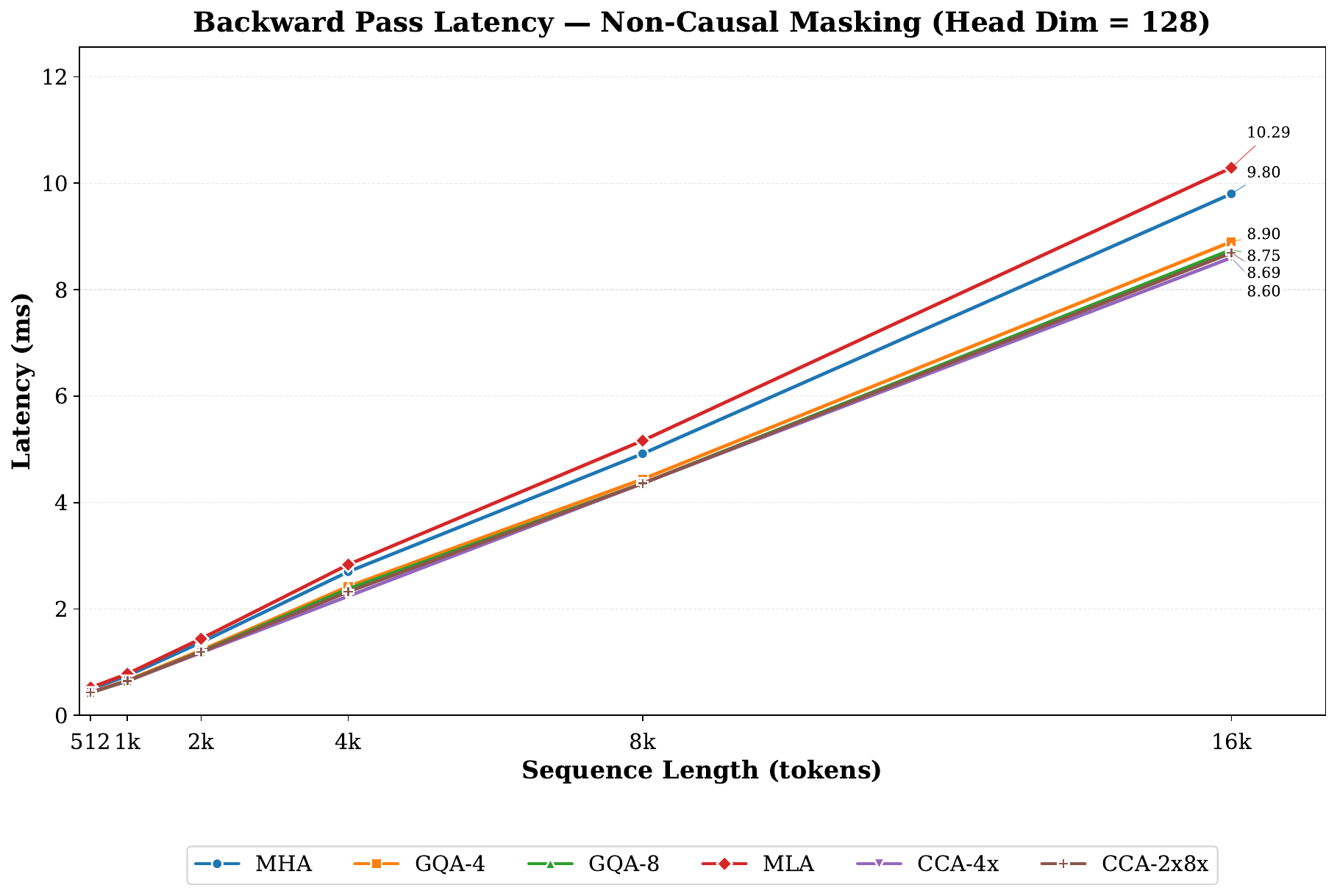}
  \caption{Backward attention latencies: Head dimension 128}
  \label{fig:bwd-128}
\end{figure*}

\begin{figure*}[htbp]
  \centering
  \includegraphics[width=0.7\textwidth]{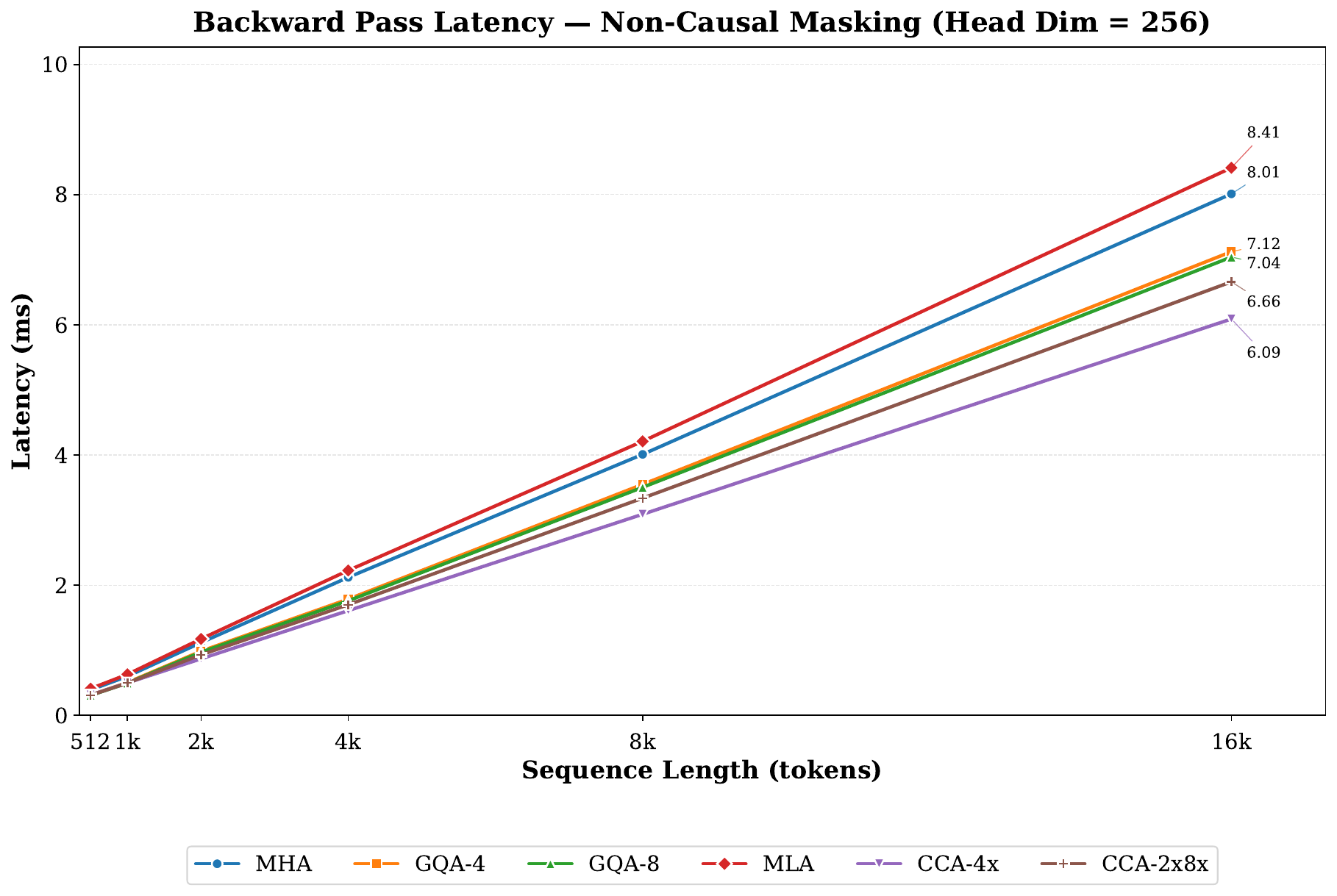}
  \caption{Backward attention latencies: Head dimension 256}
  \label{fig:bwd-256}
\end{figure*}

\par\noindent\mbox{}\vfill

\end{document}